\documentclass[10pt,twocolumn,letterpaper]{article}

\usepackage{iccv}
\usepackage{times}
\usepackage{epsfig}
\usepackage{amsmath}
\usepackage{amssymb}
\usepackage[T1]{fontenc}
\usepackage[utf8]{inputenc}
\usepackage{babel}
\usepackage[includeheadfoot,margin=2cm]{geometry}
\usepackage[font=small,labelfont=bf,tableposition=top]{caption}
\usepackage[dvipsnames]{xcolor}
\usepackage{subcaption}
\usepackage{multirow}
\usepackage{booktabs}
\usepackage{subcaption}
\usepackage[colorlinks=true, urlcolor=green, pdfborder={0 0 0},breaklinks=true,bookmarks=false]{hyperref}
\usepackage{graphicx}

 \iccvfinalcopy 

\def\iccvPaperID{12} 

\ificcvfinal\fi

\begin{document}

\title{\textsc{InFusion}: Inject and Attention Fusion for Multi Concept Zero-Shot Text-based Video Editing}

\author{Anant Khandelwal\\
Glance AI\\
{\tt\small anant.iitd.2085@gmail.com}
}

   

\twocolumn[{\maketitle
\begin{center}
    \captionsetup{type=Prompt}
    \includegraphics[height=7.5\baselineskip]{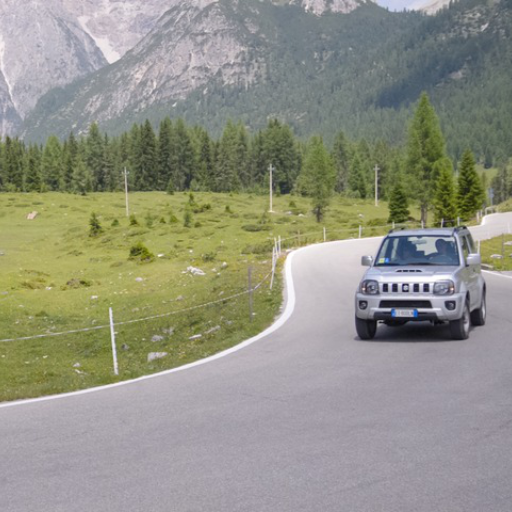} %
    \includegraphics[height=7.5\baselineskip]{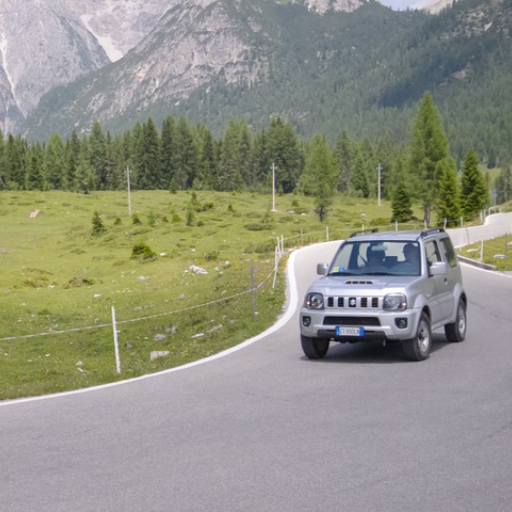} %
    \includegraphics[height=7.5\baselineskip]{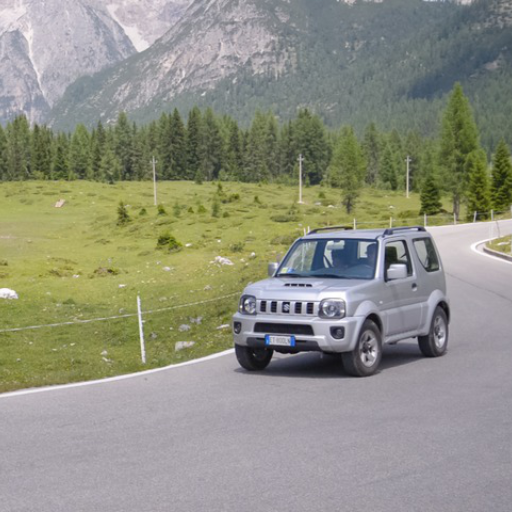} %
    \includegraphics[height=7.5\baselineskip]{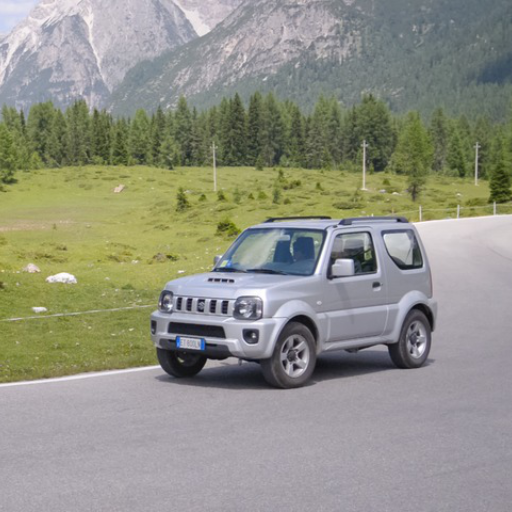} %
    \includegraphics[height=7.5\baselineskip]{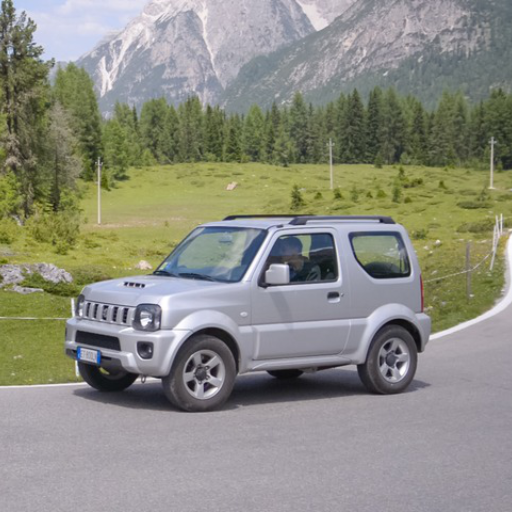}
    \vspace{-3mm}
    \captionof*{Prompt}{Source Prompt: A silver jeep driving down a curvy road in the countryside.}\label{fig:fig1}
    \captionsetup{type=Prompt2}
    \includegraphics[height=7.5\baselineskip]{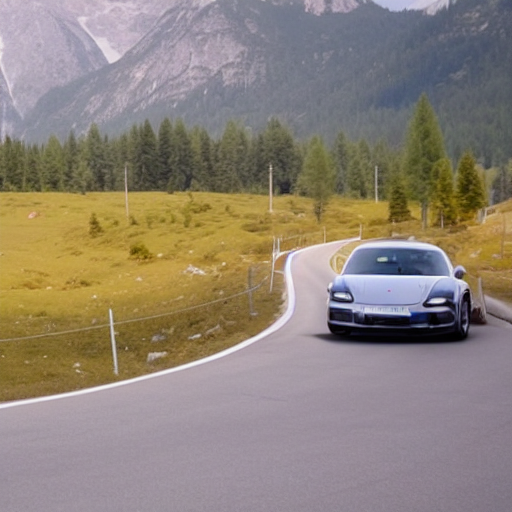} %
    \includegraphics[height=7.5\baselineskip]{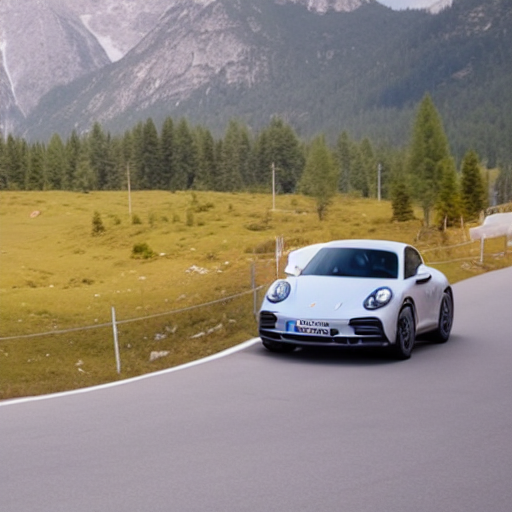} %
    \includegraphics[height=7.5\baselineskip]{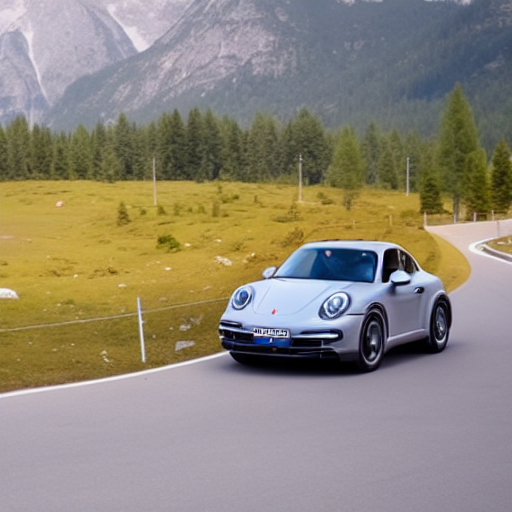} %
    \includegraphics[height=7.5\baselineskip]{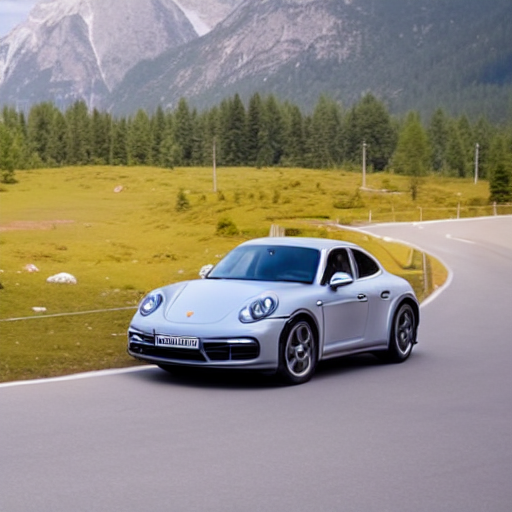} %
    \includegraphics[height=7.5\baselineskip]{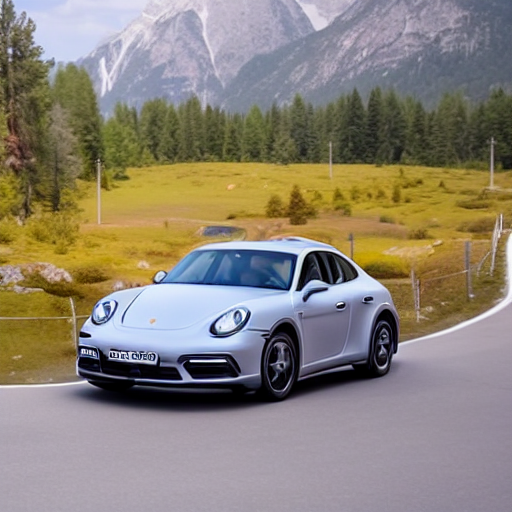}
    \vspace{-3mm}
    \captionof*{Prompt2}{Zero Shot Multi Concept Editing with Stable Diffusion v1.5 \textbf{Silver Jeep} $\rightarrow$ \textcolor{red}{Porsche car}, \textbf{countryside} $\rightarrow$ \textcolor{YellowOrange}{Landmark of autumn}}\label{fig:fig1}
    \captionsetup{type=Prompt3}
    \includegraphics[height=7.5\baselineskip]{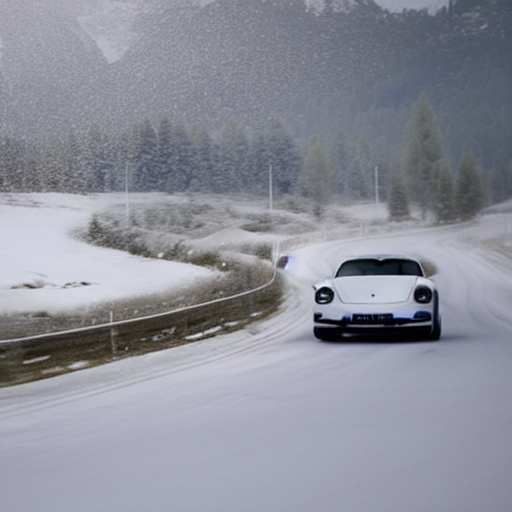} %
    \includegraphics[height=7.5\baselineskip]{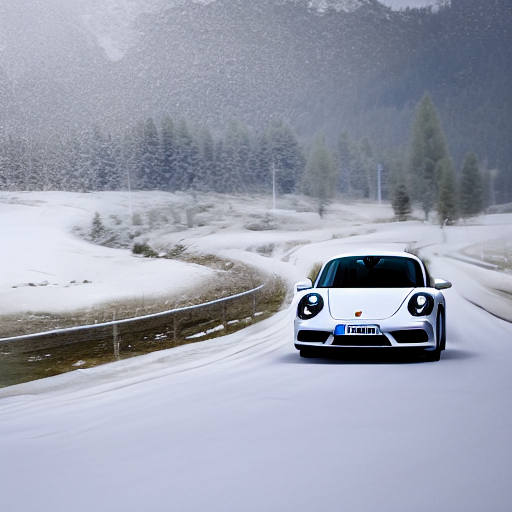} %
    \includegraphics[height=7.5\baselineskip]{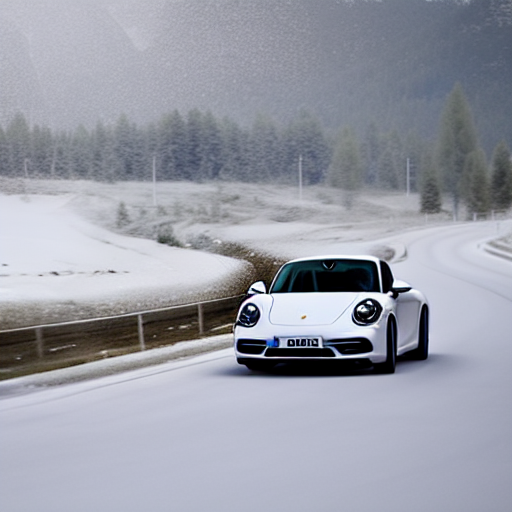} %
    \includegraphics[height=7.5\baselineskip]{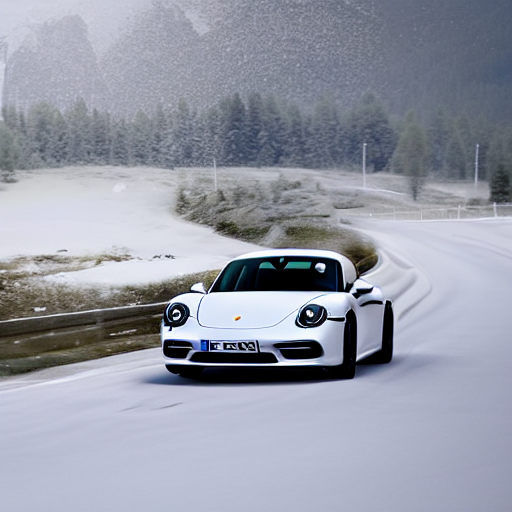} %
    \includegraphics[height=7.5\baselineskip]{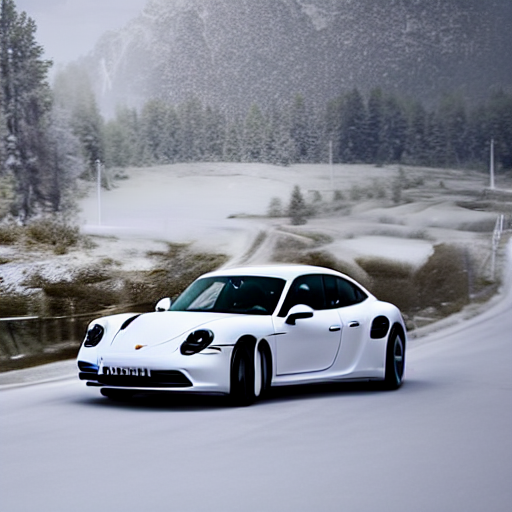}
    \vspace{-3mm}
    \captionof*{Prompt3}{Zero Shot Multi Concept Editing with Stable Diffusion v1.5 \textbf{Silver Jeep} $\rightarrow$ \textcolor{red}{Porsche car}, \textbf{countryside} $\rightarrow$ \textcolor{cyan}{snowy winter}}\label{fig:fig1}
  \end{center}
  }]
\ificcvfinal\thispagestyle{empty}\fi
\vspace{-2mm}
\begin{abstract}
\vspace{-4mm}
Large text-to-image diffusion models have achieved remarkable success in generating diverse, high-quality images. Additionally, these models have been successfully leveraged to edit input images by just changing the text prompt. But when these models are applied to videos, the main challenge is to ensure temporal consistency and coherence across frames. In this paper, we propose InFusion, a framework for zero-shot text-based video editing leveraging large pre-trained image diffusion models. Our framework specifically supports editing of multiple concepts with pixel-level control over diverse concepts mentioned in the editing prompt. Specifically, we inject the difference in features obtained with source and edit prompts from U-Net residual blocks of decoder layers. When these are combined with injected attention features, it becomes feasible to query the source contents and scale edited concepts along with the injection of unedited parts. The editing is further controlled in a fine-grained manner with mask extraction and attention fusion, which cut the edited part from the source and paste it into the denoising pipeline for the editing prompt. Our framework is a low-cost alternative to one-shot tuned models for editing since it does not require training. We demonstrated complex concept editing with a generalised image model (Stable Diffusion v1.5) using LoRA. Adaptation is compatible with all the existing image diffusion techniques. Extensive experimental results demonstrate the effectiveness of existing methods in rendering high-quality and temporally consistent videos.
\end{abstract}
\vspace{-12mm}
\section{Introduction}
With the rise in the creation and consumption of video content on social media platforms, there is a need for generalised video creation and editing tools. Despite the recent success of text-to-image diffusion models, their applicability to video is limited since per-frame editing does not produce consistent editing across all the frames. To overcome this limitation, recent research introduced three types of text-to-video diffusion: a) first solution is to train the model on large-scale video data \cite{ho2022imagen} which require lot of computing resources b) second solution is to fine-tune the image models on single video \cite{wu2022tune}  c) third solution is the zero-shot method \cite{khachatryan2023text2video, qi2023fatezero}, which requires no training, is compatible with pre-trained image diffusion models, and requires fewer computing resources. In this paper, we employ the zero-shot strategy for text-based video editing. However, the challenges associated with zero-shot methods are: 1) Temporal Consistency: Cross-Frame Continuity 2) Zero-Shot: no training or fine-tuning required 3) Flexible: compatible with off-the-shelf, pre-trained image models. In this paper, we demonstrated the use of a large-scale pre-trained text-to-image model (i.e., Stable diffusion v1.5 \cite{rombach2022high}), which contains almost all the concepts, hence can be used for any customised generation as opposed to the zero-shot method Fatezero \cite{qi2023fatezero}, which requires a one-shot tuned model for customised generation.\\
In this paper, we introduce a novel zero-shot framework for text guided video editing with fine grained control over multiple concepts. Our framework, \textsc{InFusion}, consists of two parts \textsc{Inject} and \textsc{Attention Fusion}. In the first part, we inject features from residual block in decoder layers and attention features (obtained from source prompt $P_s$) into the denoising pipeline for editing prompt $P_e$. This injection step highlights the target concepts since we injected the difference between residual block features for ($P_s$, $P_e$) and combined them with attention injection (keys and values) to query the source contents, keeping the unedited concepts as they are and scaling up the edit concepts in the edit pipeline while scaling down the removed concepts from the source pipeline. In the second part, we fuse the attention for edited and unedited concepts using the mask extraction obtained from cross-attention maps for $P_e$ and $P_s$, respectively. The fused attention preserves the source content with editing concepts. Additionally, we mix the cross-attention from the source and edited prompts to contain the unedited and edited concepts, respectively. To summarize, our main contributions are as follows: 
\begin{itemize}
    \item A novel zero-shot framework capable of editing multiple concepts with finer details with a single editing pipeline. It achieves the best temporal consistency and generates coherent edited videos, with no training involved either for the generalised image diffusion model or for edited video generation.
    \item \textsc{Inject} for fine-grained control over editing concepts and \textsc{Attention Fusion} to cut the edited part and paste the unedited part from source attention.
    \item Experimental results demonstrate the flexible structure, shape, colour, and style of editing with a temporally coherent generation of edited videos.
\end{itemize}
\section{Related Work}
Large-scale zero-shot methods for text-based image editing triggered interest in videos as well. Recent developments introduced video editing methods, namely, Tune-A-Video \cite{wu2022tune} is a one-shot method that inflates an image diffusion model into a video model with cross-attention and generates edited video by fine-tuning on a single video. Other methods based on the same idea are Edit-A-Video \cite{shin2023edit}, VideoP2P \cite{liu2023video} and vid2vid-zero \cite{wang2023zero} which uses Null-text inversion \cite{mokady2023null}for preserving unedited regions. However, all these methods require fine-tuning of the pre-trained model over the input video. Following these zero-shot methods are introduced, namely, FateZero \cite{qi2023fatezero} proposed attention blending using features before and after editing, Text2Video-Zero \cite{khachatryan2023text2video} denoise the latent directly to motions, Pix2Video \cite{ceylan2023pix2video} matches the current frame with the previous frame in latent space. All the mentioned zero-shot methods largely rely on manipulation with cross-attention maps for early-step latent fusion to improve temporal consistency. However, as we demonstrate, these methods are effective in editing high-level styles and shapes but less effective in manipulating concepts at fine-grained levels. Our method does the editing at finer levels using feature injection, which acts at pixel level. Over and above, we apply attention feature injection and fusion to control over the concepts mentioned for editing.
\section{Preliminary}
\label{prem}
\textbf{Latent Diffusion Models}: Diffusion models \cite{rombach2022high, ho2020denoising, nichol2021improved, song2020denoising} are probabilistic generative models that can generate the desired image from an initialised Gaussian noise image $x_T \sim \mathcal{N}(0, \mathbf{I})$ by progressively removing the noise at step ranging from $T$ to $0$. In general, the foundation of diffusion models is based on two complementary random processes i.e. \textit{forward} and \textit{backward}. During \textit{forward process} or \textit{inversion} the noise is added at each step from $0$ to $T$ to clean image $x_0$ defined as:
\begin{equation}
    x_t = \sqrt{\alpha_t} \cdot x_0 + \sqrt{1 - \alpha_t} \cdot z
\end{equation}
where $z \sim \mathcal{N}(0, \mathbf{I})$ and {$\alpha_t$} are the noise schedule. \\
The \textit{backward process} or \textit{reconstruction} is aimed at progressively denoising the image $x_T$, where at each step $t$ the cleaner version of image is obtained than the previous step $t+1$, and finally to cleaned image at $0$. This is achieved by a neural network $\epsilon_{\theta}(x_t, t)$, which predicts the added noise $z$ at each step. Once trained, this is applied at each backward step which consists of applying $\epsilon_{\theta}$ to the current $x_t$, and adding a Gaussian noise perturbation to obtain a cleaner $x_{t-1}$, defined as:
\begin{align}
    p_{\theta}(x_{t-1} | x_t) = \mathcal{N}(x_{t-1}; \mu_{\theta}(x_t, t), \sigma_t), \\
    \mu_{\theta}(x_t, t) = \frac{1}{\sqrt{\alpha_t}}(x_t - \epsilon \frac{\sqrt{1 - \alpha_t}}{\sqrt{1 - \bar{\alpha}_t}}), 
\end{align}
where $\bar{\alpha}_t = \prod_{i}^{t}\alpha_t$, and $\epsilon$ is the predicted noise. Neural network $\epsilon_{\theta}$ is trained using the mean squared error given as: 
\begin{equation}
    \mathcal{L} = \mathbb{E}_{x_0, \epsilon, t}(||\epsilon - \epsilon_{\theta}(x_t, t)||)
\end{equation}
Diffusion models are evolving very fast and have been integrated and trained to generate images conditioned on multiple types of guiding signals, denoted as $y$ in $\epsilon_{\theta}(x_t, y, t)$ i.e. another image \cite{saharia2022palette}, text \cite{kim2022diffusionclip, nichol2021glide, ramesh2022hierarchical, rombach2022high} or class label \cite{ho2022cascaded}. In this work, we leveraged the pre-trained text-conditioned Latent Diffusion Model (LDM), a.k.a. Stable Diffusion \cite{rombach2022high}, which performs the diffusion-denoising process in the latent space of the pre-trained image auto-encoder network. The structure of the denoising backbone $\epsilon_{\theta}$ is realized as a time-conditional U-Net \cite{ronneberger2015u} conditioned on the guiding text prompt $P$.\\
\textbf{Self-Attention and Cross-Attention}: Layers of denoising U-Net consists of a residual block \cite{he2016deep}, a self-attention block and a cross-attention block \cite{vaswani2017attention}. At the denoising step $t$, the residual block convolves features from previous layer $\phi_{t}^{l-1}$ to produce the intermediate features $f_t^l$ at the layer $l$. In self-attention block these intermediate features are projected to produce the queries $q_t^l$, keys $k_t^l$ and values $v_t^l$. The output feature of self-attention is then given as:
\begin{equation}
    \hat{f}_t^{l} = A_t^{l}v_t^{l}, \textrm{    where   }  A = Softmax(q_t^l{k_t^l}^\mathbf{T}) 
    \label{att}
\end{equation}
Finally, the textual prompt $P$ features are projected into keys and values, which are queried by self-attended spatial features, which when plugged into the attention equation \ref{att} will compute the features at the output of the cross-attention block. These attention maps in the stable diffusion collectively contain the rich information of structure, shape, and layout present in the spatial features obtained from residual blocks. Cross-attention maps the spatial pixels to the input text prompt and allows the editing \cite{hertz2022prompt} of multiple granular objects that are present in the source video. Meanwhile, the features in self-attention layers are employed in a plug-and-play manner \cite{tumanyan2023plug} to retain the structure/ layout/ shape of un-edited objects and facilitate style editing over them with the edited prompt. Collectively, in this work, we leveraged the combination of cross-attention and self-attention maps to perform consistent video synthesis with delicate handling of multi-concept editing in a zero-shot manner that can preserve and retain the layout and structure of edited and un-edited parts in a prompt, respectively.

\begin{figure*}[hbt!]
    \centering
    \includegraphics[width=\textwidth]{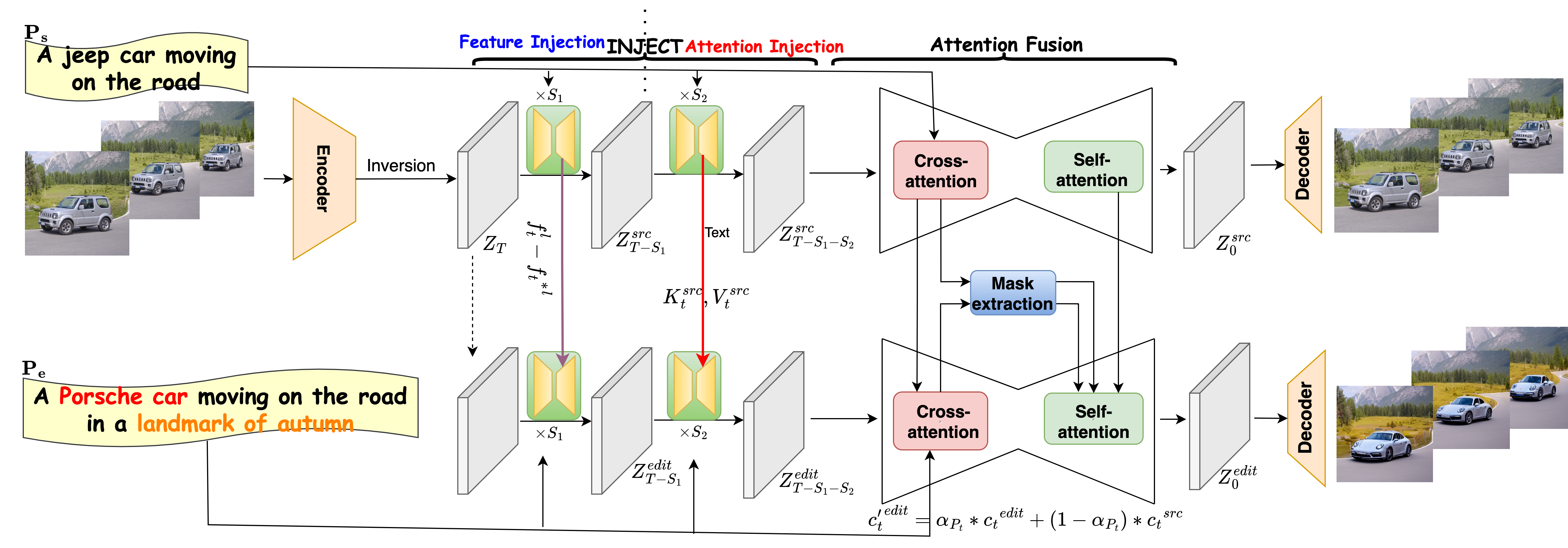}
    \caption{\textsc{\textbf{InFusion}}: Leveraging a pre-trained text-to-image model for video editing ensures temporal consistency and editing accuracy with Inject and Attention Fusion. The denoising pipeline for source prompt $P_S$ generates the decoder latent from U-Net and attention features from source video, which are injected into the denoising pipeline (initialised with inverted source latent $z_T$) for edit prompt $P_e$.}
    \label{fig:arch}
    \vspace{-6mm}
\end{figure*}
\section{InFusion}

In this section we present \textit{InFusion}, a framework designed to do zero shot text based video editing of multiple concepts and ensuring temporal consistency between frames of the edited video. Formally, given the source input video $X_0 = \{x_0\}_{i=1}^N$ with $N$ frames, source prompt $P_s$ and the target prompt $P_e$, the goal of text driven video editing is to generate a video $Y_0 = \{y_0\}_{i=1}^N$ which aligns with prompt $P_e$, faithfully preserves the unedited content of source video $X_0$ and maintains the temporal consistency between frames. Our framework is built upon Stable Diffusion v1.5 \cite{rombach2022high} a pre-trained and fixed text-to-image LDM model denoted by $\epsilon_{\theta}(x_t, P, t)$ where $P$ is the given prompt. This model is based on the U-Net architecture with $T$ time-steps denoising as shown in Figure \ref{fig:arch} and discussed in Section \ref{prem}. However, this model can generate the frames as per the given prompt but to ensure temporal consistency between frames and retaining the unedited contents of source video we made several modifications to the pipeline. \\
Our key finding is that the fine grained control over the generated structure is achieved by highlighting each concept using the a) edit directions obtained from difference of spatial features from source and edit prompts b) accurate mask extraction from source and edited cross-attention maps for fine grained control over the structure of edited shape c) retain the unedited structure by combining cross-attention maps from source and edit prompts for source and edited parts respectively.
\begin{figure*}
\centering
    \raisebox{0.5in}{\rotatebox[origin=t]{90}{\textbf{Input}}} %
    \includegraphics[ height=7\baselineskip]{00001.png}%
    \includegraphics[ height=7\baselineskip]{00002.png}%
    \includegraphics[ height=7\baselineskip]{00003.png}%
    \includegraphics[ height=7\baselineskip]{00004.png}%
    \includegraphics[ height=7\baselineskip]{00005.png}\\
    \raisebox{0.5in}{\rotatebox[origin=t]{90}{\textbf{layer 4}}} %
    \includegraphics[ height=7\baselineskip]{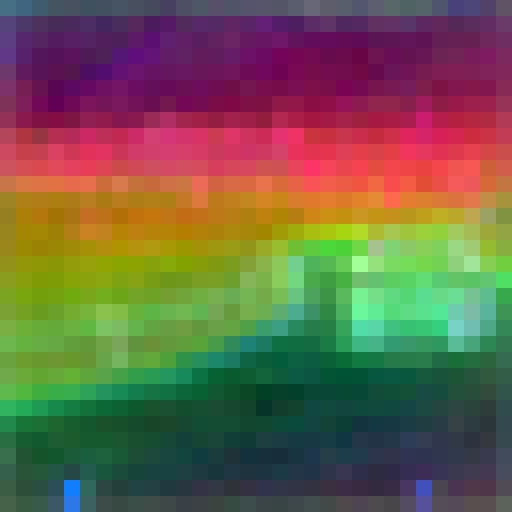}%
    \includegraphics[ height=7\baselineskip]{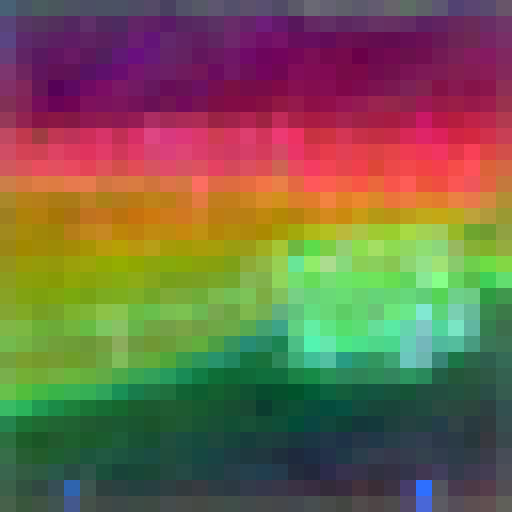}%
    \includegraphics[ height=7\baselineskip]{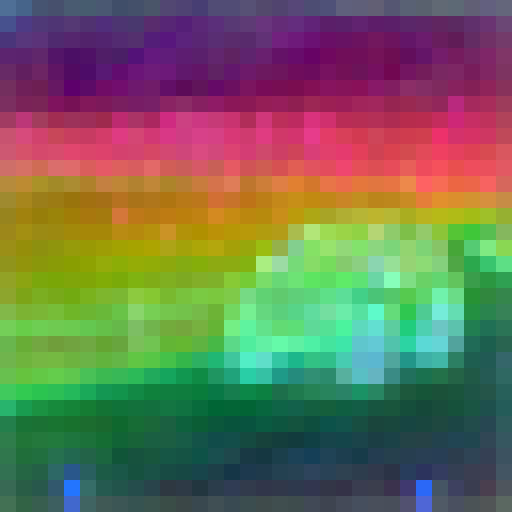}%
    \includegraphics[ height=7\baselineskip]{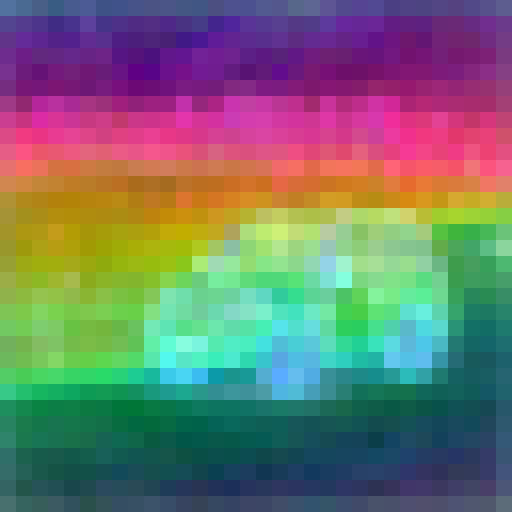}%
    \includegraphics[ height=7\baselineskip]{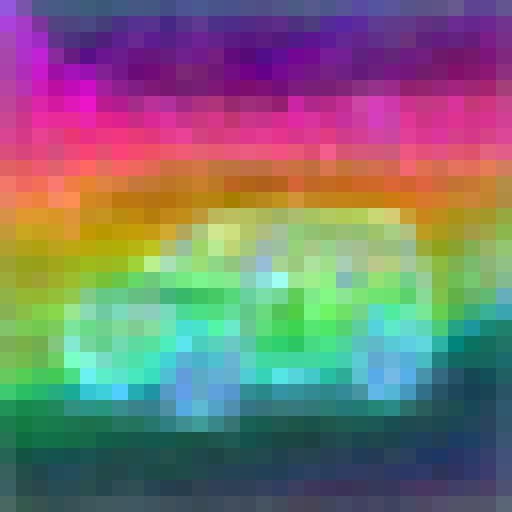}\\
    \raisebox{0.5in}{\rotatebox[origin=t]{90}{\textbf{layer 7}}} %
    \includegraphics[ height=7\baselineskip]{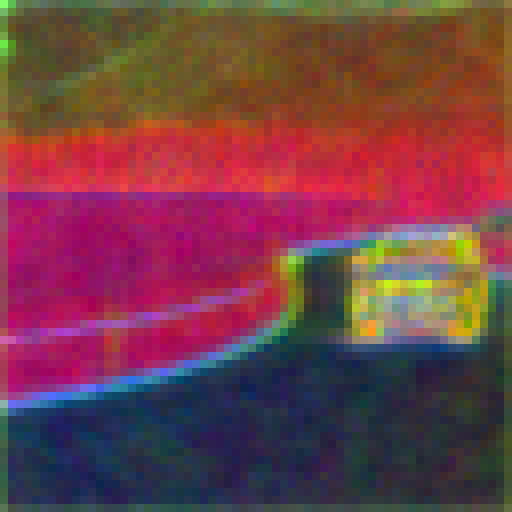}%
    \includegraphics[ height=7\baselineskip]{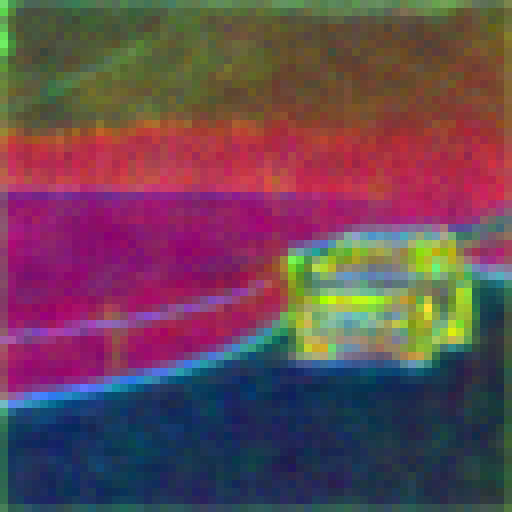}%
    \includegraphics[ height=7\baselineskip]{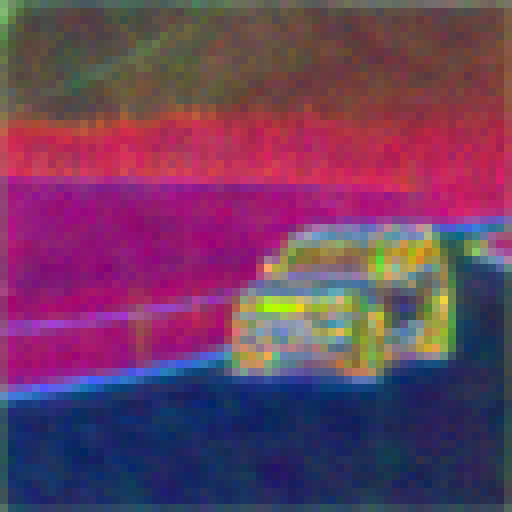}%
    \includegraphics[ height=7\baselineskip]{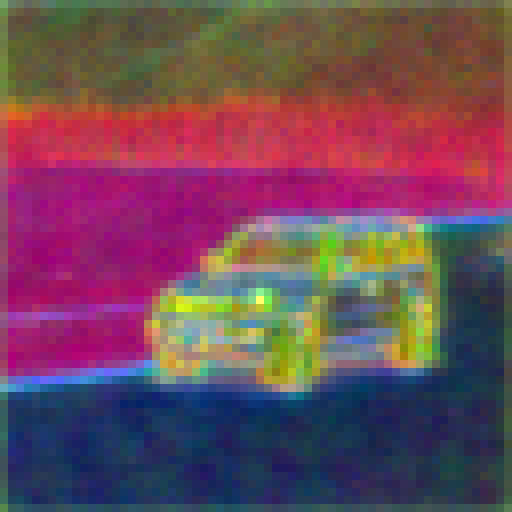}%
    \includegraphics[ height=7\baselineskip]{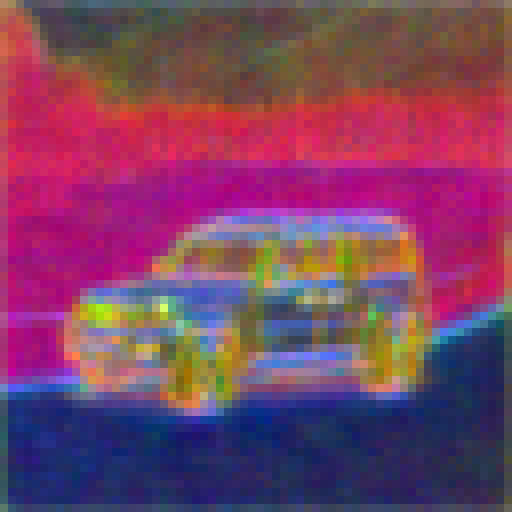}\\
    \raisebox{0.5in}{\rotatebox[origin=t]{90}{\textbf{layer 11}}} %
    \includegraphics[ height=7\baselineskip]{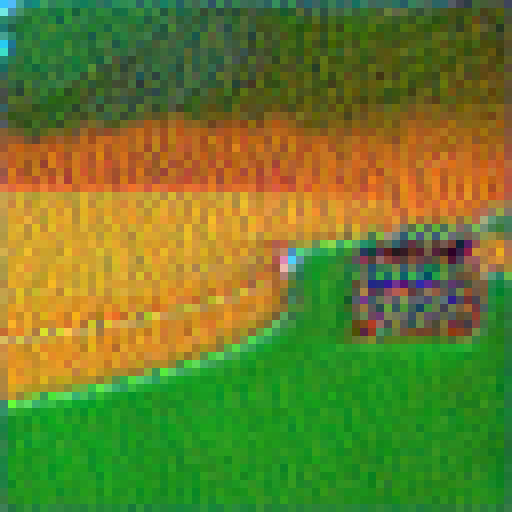}%
    \includegraphics[ height=7\baselineskip]{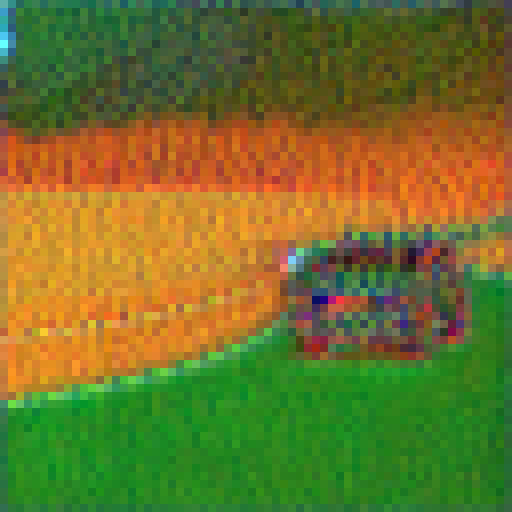}%
    \includegraphics[ height=7\baselineskip]{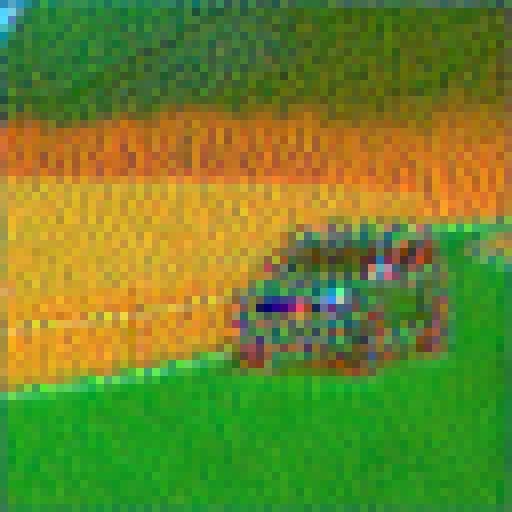}%
    \includegraphics[ height=7\baselineskip]{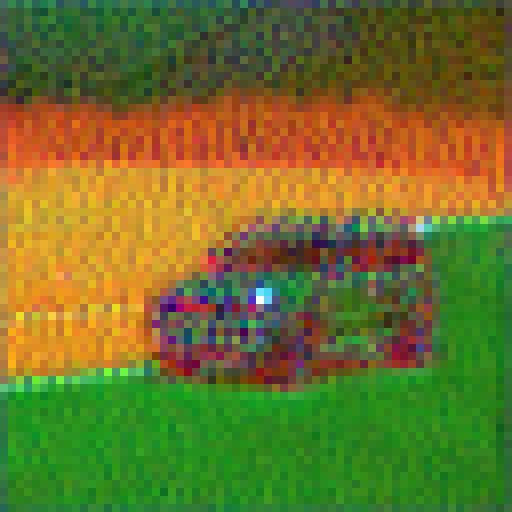}%
    \includegraphics[ height=7\baselineskip]{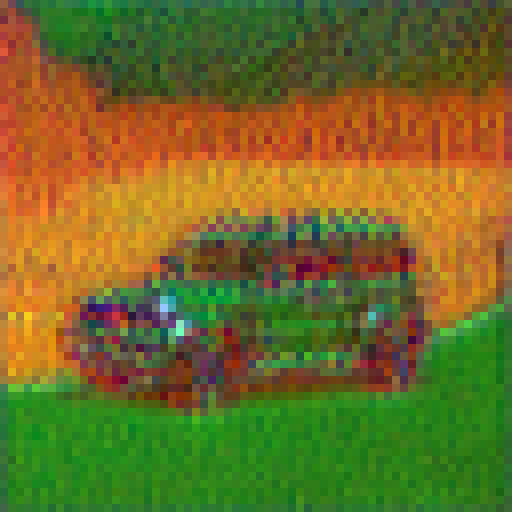}\\
    \raisebox{0.5in}{\rotatebox[origin=t]{90}{$f_t^4 - {{f_t}^{*}}^{4}$}} %
    \includegraphics[ height=7\baselineskip]{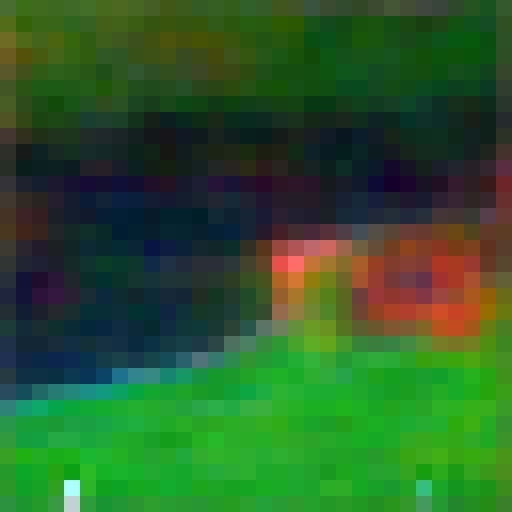}%
    \includegraphics[ height=7\baselineskip]{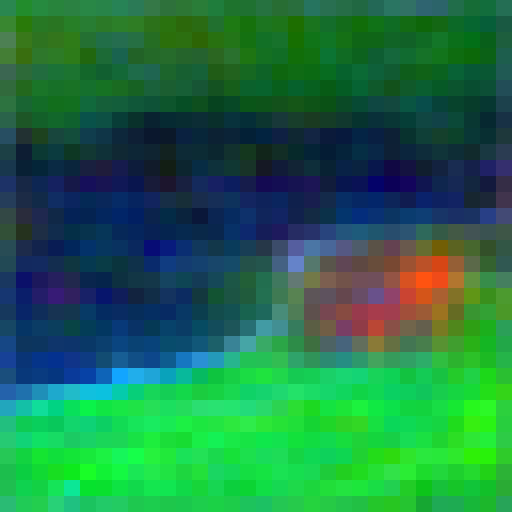}%
    \includegraphics[ height=7\baselineskip]{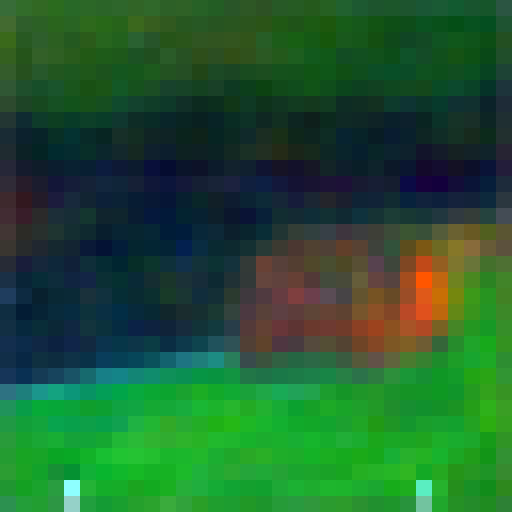}%
    \includegraphics[ height=7\baselineskip]{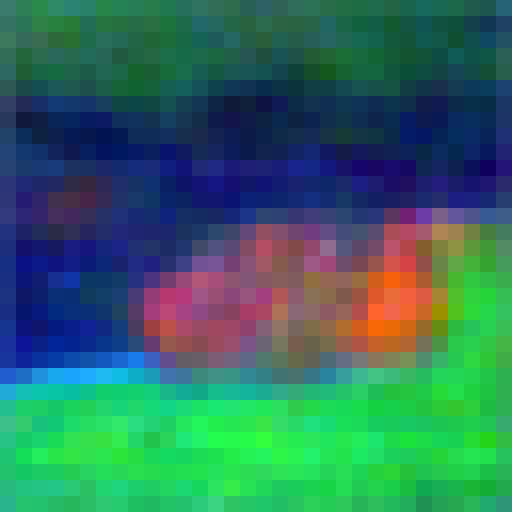}%
    \includegraphics[ height=7\baselineskip]{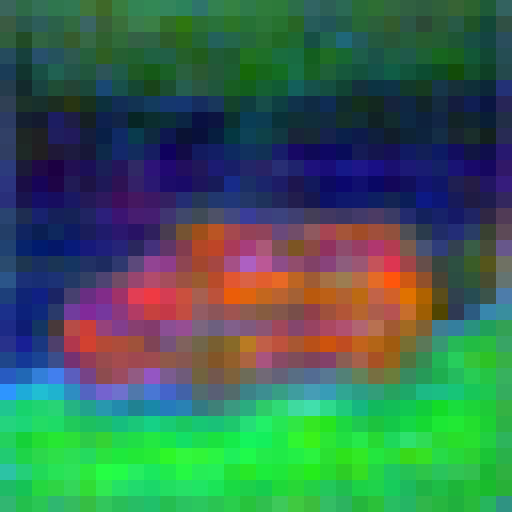}
    \caption{Visualising Top-3 principal components of diffusion features (spatial features) obtained from the decoder of U-Net at different layers.}
    \label{fig:spatial}
    \vspace{-4mm}
\end{figure*}
\subsection{\textsc{Inject}}
\textbf{Spatial Features}: Spatial features in text-to-image generation methods govern the basic part of specifying the structure/shape/pose/scene layout. Even if the prompt is descriptive, like "\textit{a Porsche car driving down a curvy road in the countryside}" or "\textit{a cat jumping over the bed}" the model can generate different images under different initial noise $x_T$. We hypothesise that in text-based editing, the structure/pose can be controlled in a fine-grained manner using the spatial features, and this hypothesis is motivated by the analysis in \cite{baranchuk2021label, tumanyan2023plug}, which demonstrated the semantic segments obtained from spatial features. To further investigate this fact, we did a PCA analysis, as shown in Figure \ref{fig:spatial}. Specifically, for each input image, we extract the features $f_t^l$ from each layer in decoder of  $\epsilon_{\theta}$ at each time-step and compute the first three principal components as displayed in Figure \ref{fig:spatial} for layers 4,7 and 11. As seen, in the coarsest layer (layer 4), a crude blob of Jeep structure is visible, but in layers 7 and 11, the Jeep structure is clearly visible. Interestingly, the colour of the similar object (irrespective of its pose) is same across all the frames at each layer. In text-based video editing, we have to retain the source layout, and hence we choose to inject the source features while editing with the prompt $P_e$, but while injecting, we have to edit the structure of some objects, so we choose to inject the source features in coarse layers only, since features at higher layers gradually capture more fine-grained information based on the features at coarse layers, and since these features eventually contribute to the error predicted by the U-Net, the tendency will be more towards decreasing the error at finer levels.\\
\textbf{Feature Injection and Edit Direction}: We now discuss the translation of the given source ($x_0$, $P_s$) to edited video $y_0$ from the edited prompt $P_e$. First, the source video is inverted using DDIM\cite{song2020denoising} to noise denoted as $z_T$. Given the target prompt $P_e$, the generation of edited video $y_0$ is performed using the same initial noise $z_T$ as shown in Figure \ref{fig:arch}. At each step $t$ of the backward process from initial noise $z_T$ for source prompt, the guidance features $\{f_t^l\}$ are collected at each layer from the denoising step $z_{t-1} = \epsilon_{\theta}(x_t, P_s, t)$. We then inject these source guidance features $\{f_t^l\}$ during the denoising steps of $y_t$ from the target prompt $P_e$. Specifically, we replace the resulting features $\{{f_t^{*}}^{l}\}$ given as follows:
\begin{equation}
    z_{t-1}^{*} = \epsilon_{\theta}(y_t, P_e, t; \{f_t^l - {f_t^{*}}^{l}\}) \textrm{   where   }  y_T = z_T
\end{equation}
The edited prompt "\textit{a Porsche car driving down a curvy road in a landmark of autumn}" contains the following edited concepts: a) "\textit{\textbf{silver jeep}} $\rightarrow$ \textit{\textcolor{red}{Porsche car}}" b) "\textit{\textbf{countryside}} $\rightarrow$ \textit{\textcolor{YellowOrange}{landmark of autumn}}". As shown in Figure \ref{fig:spatial} the spatial features $\{f_t^l - {f_t^{*}}^{l}\}$ at \textit{layer 4}, the jeep structure which was visible in $f_t^l$ for source prompt $P_s$ is now not clearly visible (looks moving towards car structure in some frames), and the colour of the "\textit{countryside}" is also changed, depicting that it is moving from source concepts to edited concepts. Hence, we instead injected the  $\{f_t^l - {f_t^{*}}^{l}\}$ features since we want to move from source concepts to edited concepts as mentioned in a) and b) rather than retaining them. However, to reflect the complete structure change for edited concepts, feature injection is not enough since it can only give the edit directions (after a few steps $S_1$ as shown in Figure \ref{fig:arch}), these are leveraged in controlling the self-attention to cut the edited concepts from the source structure and paste the remaining part without any modification.\\
\begin{figure}
    \centering
    \begin{subfigure}[b]{\linewidth}
        \includegraphics[width=\linewidth]{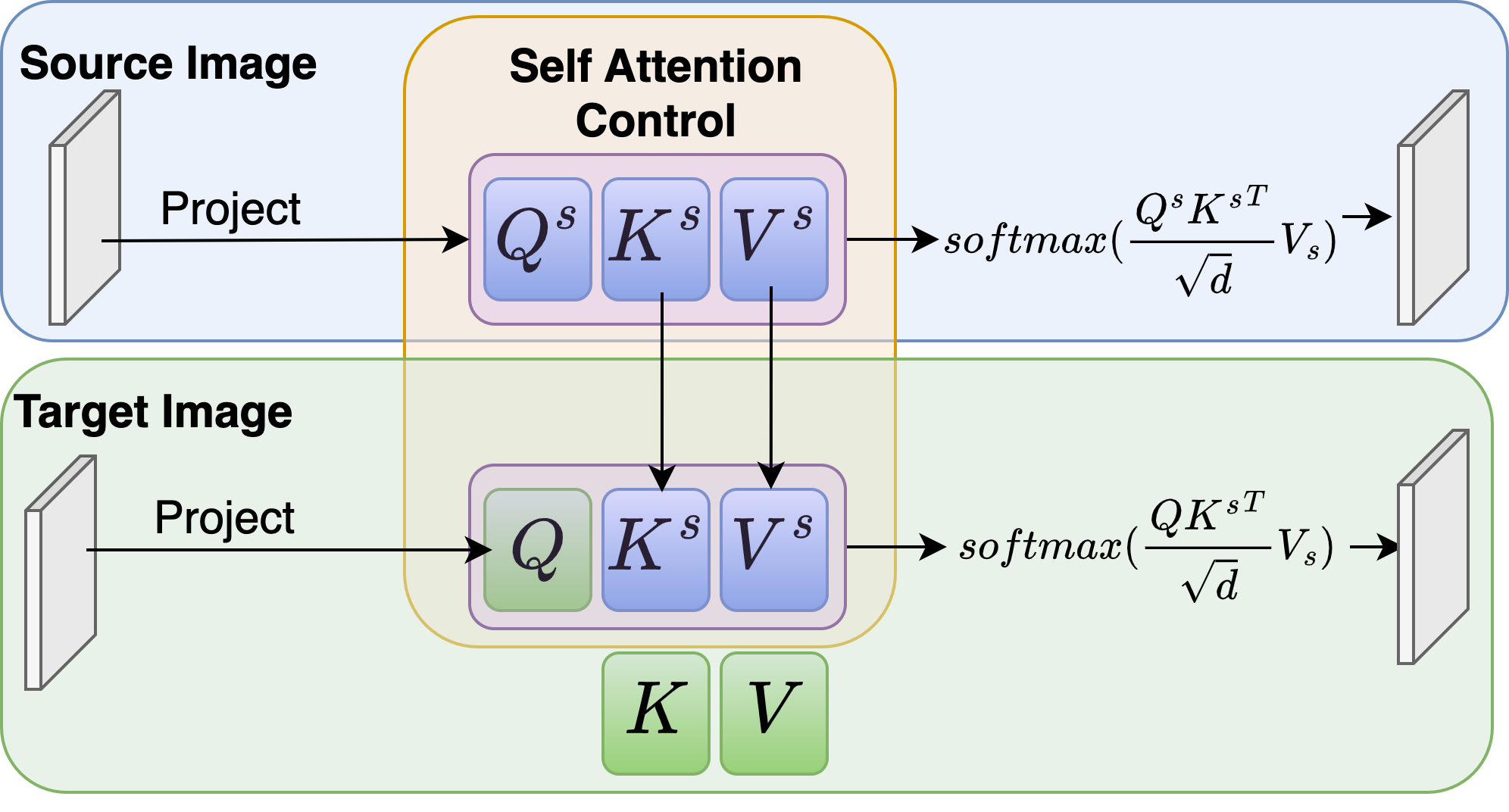}
        \caption{Self-Attention Control}
        \label{control}
    \end{subfigure}
    
    \begin{subfigure}[b]{\linewidth}
        \includegraphics[width=\linewidth]{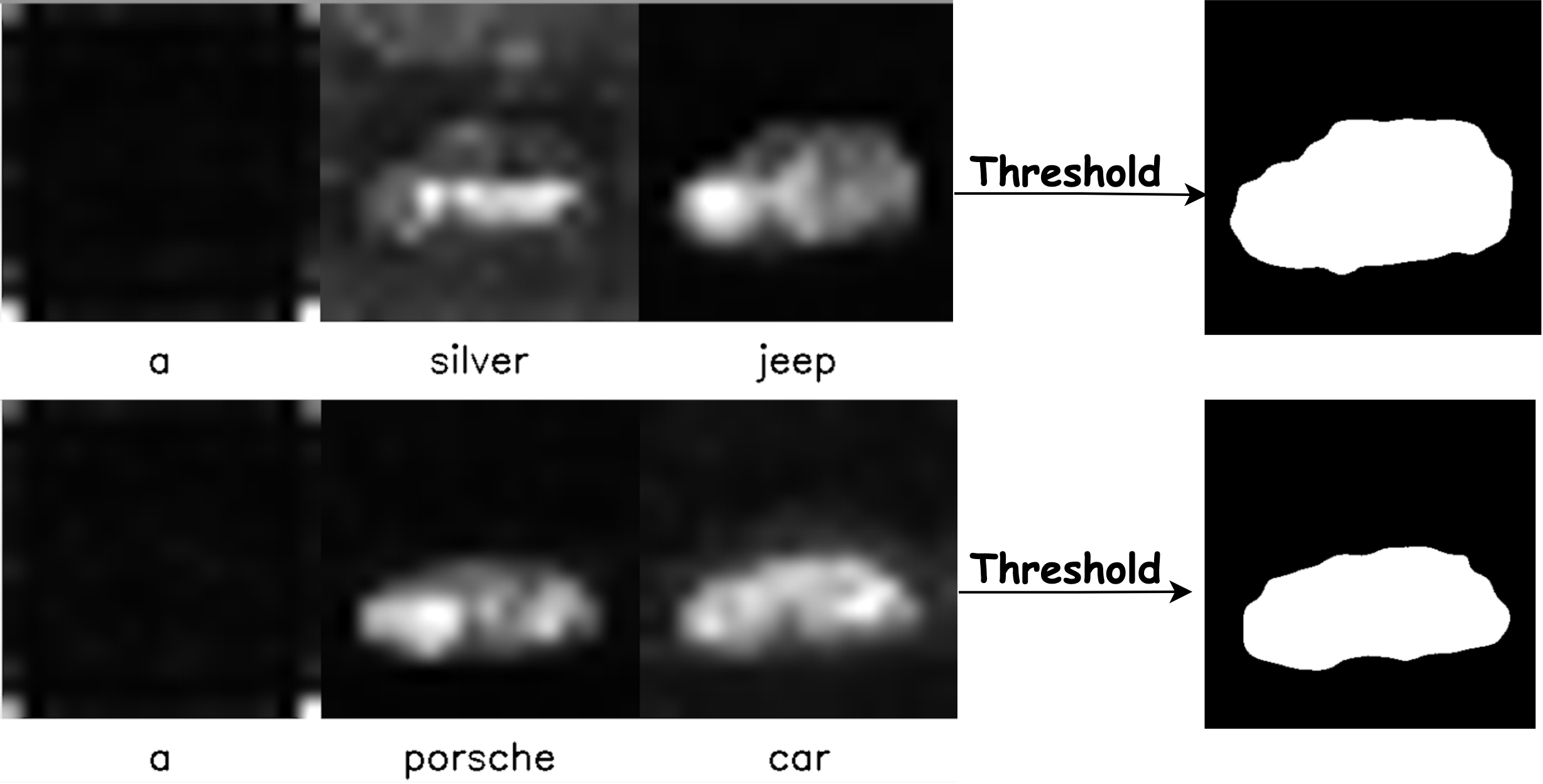}
        \caption{Mask Extraction from cross-attention maps}
        \label{mask}
    \end{subfigure}
    \caption{a) Self-attention control to query contents from source image in decoder part of U-Net b) Mask extraction strategy}
\vspace{-4mm}
\end{figure}
\textbf{Self-Attention Control}: Figure \ref{control} depicts the mechanism of self-attention control, where the keys ${K_t^{s}}^{l}$ and values ${{V_{t}}^{s}}^{l}$ from source (collected during denoising step of $x_t$) are injected during denoising step of $y_t$. Specifically, as shown in Figure \ref{control} the queries $Q$ obtained from the injected spatial features $\{f_t^l - {f_t^{*}}^{l}\}$ downscale the affinities for the edited concepts (a and b as discussed in feature injection) due to the structure changed for those concepts in feature injection and hence less similarity for those concepts with keys ${K_t^{s}}^{l}$, less probability for the edited concepts, which scale down the edited concepts in ${{V_{t}}^{s}}^{l}$. This ensures the self-attention needed to control the propagation of edited parts. However, the cross-attention maps, which correlate to the target prompt using keys and values, make the edit concepts slowly integrate into the source layout. Once the source layout with edited concepts is brewed in the diffusion process (after $S_2$ steps as shown in Figure \ref{fig:arch}) we perform the mask-guided mixing of self-attention and cross-attention from source and target prompt. The \textsc{Inject} operation is defined as follows:
\begin{equation}
    z_{t-1}^{*} := \left\{\begin{matrix}
 \epsilon_{\theta}(y_t, P_e, t; \{f_t^l - {f_t^{*}}^{l}\}), t \in [0, S_1), l < L
\\
\epsilon_{\theta}(y_t, P_e, t; \{{K_t^{s}}^{l}, {{V_{t}}^{s}}^{l}\}),  t \in [S_1, S_2], \forall l 
\end{matrix}\right.
\end{equation}
\subsection{\textsc{Attention Fusion}}
\textbf{Cut and Paste Self-Attention}: We observed that synthesised videos using the \textsc{Inject} operation faithfully generate the source layout with noise in the edited concepts like overlapping parts of "\textit{jeep}" and "\textit{Porsche car}". Hence, we propose the use of mask-guided editing of self-attention maps for faithful reconstruction of edited concepts without any overlapping parts from the source for those edited concepts. Inspired by the previous works \cite{hertz2022prompt, tang2022daam, cao2023masactrl}, it is revealed that cross-attention maps correlate to the target prompt and hence can be used to inject the edited concepts from the attention maps obtained from the edit prompt while keeping the source layout. Specifically, at step $t$, we store all the self-attention and cross-attention maps from source prompt $P_s$ and edit prompt $P_e$ during denoising steps using fixed backbone U-Net with \textsc{Inject} in place for the edit prompt $P_e$. Then we average the source and edit cross-attention maps for edited words in $P_e$ across all the heads and layers with spatial resolution $16 \times 16$, the resulting maps are denoted as ${A_c^{t}}^{src} \in R^{16 \times 16 \times N}$ (N is the number of tokens edited) and similarly in target attention for edited words denoted as ${A_c^{t}}^{edit} \in R^{16 \times 16 \times N}$. We then calculate the masks (shown in Fig. \ref{mask}) by thresholding the max pooled maps of both ${A_c^{t}}^{src}$ and ${A_c^{t}}^{edit}$ denoted as $M_s$ and $M_e$ respectively. This captures only the foreground edited objects in the binary mask. The resulting mask-guided self-attention is given as:
\begin{align}
    s_t^{fused} &= M_e * s_t^{edit} + (1 - M_s) * s_t^{src}\\
    s_t^{fused} &= fill(equal(s_t^{fused}, 0), s_t^{edit})
    \label{selfusion}
\end{align}
which denotes that we take the foreground edited objects from target attention maps only, which is "\textit{cut}" and "\textit{paste}" the remaining background from source using $1 - M_s$.\\
\textbf{Cross-Attention Fusion}: Similarly, the final fused cross-attention is obtained by taking the target cross-attention maps $c_t^{edit}$ for all heads and all layers for the edited words and taking the source attention $c_t^{src}$ for unedited words. Mathematically, the fused cross-attention is given as:
\begin{equation}
    c_t^{fused} = \alpha_w * c_t^{edit} + (1 - \alpha_w) * c_t^{src} 
\end{equation}
where $\alpha_w$ is the $1/0$ array indicating $1$'s where the word index is edited and $0$ for source words in the edit prompt $P_e$ as compared to source prompt $P_s$. 
\subsection{\textsc{Spatio-Temporal Attention}}
Inspired by the previous works \cite{wu2022tune, qi2023fatezero, zhao2023controlvideo} we also leveraged the spatio-temporal attention for consistent video synthesis for the edit prompt. Since it has been observed that spatial features are the basic foundation of structure in the synthesised video, we initialised the weights of temporal attention with weights of spatial self-attention. Specifically, we integrated the key frame attention into the spatial self-attention to align all the frames with the key frame. Formally, let $z^i$ and $z^k$ denote the embedding of the $i$-th frame and key frame, respectively. The modified spatial attention is given as:
\begin{equation}
    Q = W^Qz^i,  K = W^K[z^i; z^k], V = W^V[z^i; z^k]
\end{equation}
where [;] denotes concatenation, $W^Q, W^K, W^V$ are the projection matrices of the pre-trained model. Among the possible key-frame choices: a) $k = \textrm{round}(\frac{N_F}{2}$ where $N_F$ is the total number of frames) b) $k = i-1$ c) $k = i+1$. We find that there are no significant differences in using any of the above choices, and hence, for ease, we choose to take k=i-1 as the key frame. The resulting spatio-temporal attention map is represented as $s_t \in R^{hw \times 2hw}$ where $2$ denotes the spatio-temporal correspondence considered in calculating attention at a given time step. Overall, this concludes the end-to-end zero shot editing with the general-purpose diffusion model, which requires no fine tuning for editing the concepts that are already present in the trained model.
\section{Experimental Results}
\subsection{Implementation Details} 
All the experiments are conducted on one NVIDIA Tesla A100 40GB GPU. For zero-shot text-based video editing, we use the trained Stable Diffusion 1.5 \cite{rombach2022high} as the base text-to-image model, which has been converted to a video editing model by incorporating spatio-temporal attention along with \textsc{Inject} and \textsc{Attention Fusion} as shown in Figure \ref{fig:arch}. Throughout all experiments, the DDIM sampler was used with $T=50$ steps and 7.5 as the classifier guidance for the video editing pipeline. Out of 50 steps, the total number of steps for \textsc{Inject} operation is 12, of which 6 are for feature injection and the rest 6 are for self-attention injection, denoted as $S_1$ and $S_2$ in Figure \ref{fig:arch} respectively. We explained above the choice of feature injection for coarse layers only, hence L = 6 in our case. The mask threshold is set to 0.3 for the case of editing "\textit{Porsche car}", but it is subject to change from case to case. Rest of the steps are consumed by \textsc{Attention Fusion} operation. Following existing works \cite{qi2023fatezero, zhao2023controlvideo, bar2022text2live, esser2023structure}, we evaluated our method on videos from DAVIS\cite{pont20172017} dataset. The source prompt $P_s$ for these videos is obtained using the caption model \cite{li2022blip}. We develop the edit prompt $P_e$ by adding or replacing some words. 
\begin{figure}
\centering
\begin{subfigure}[b]{0.25\linewidth}
   \includegraphics[width=\linewidth]{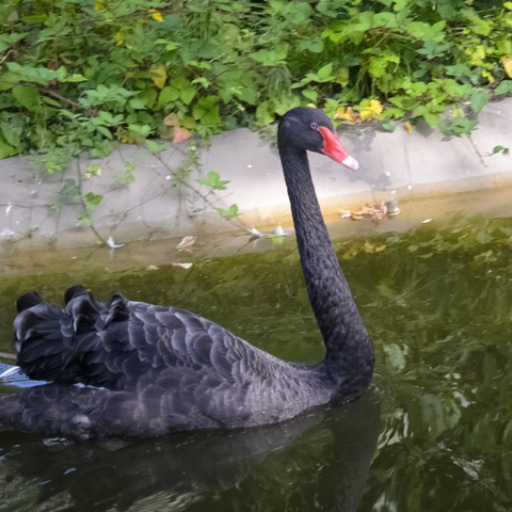}\\
    \includegraphics[width=\linewidth]{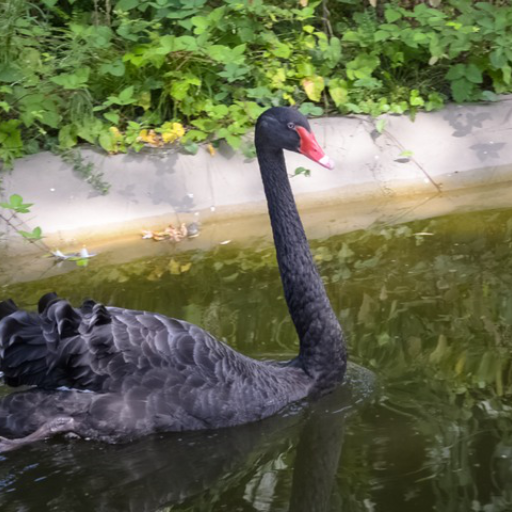}
   \caption{Input}
\end{subfigure}%
\begin{subfigure}[b]{0.25\linewidth}
        \includegraphics[width=\linewidth]{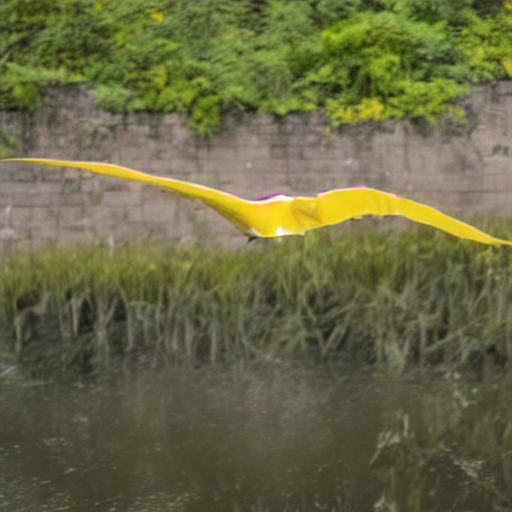}\\
        \includegraphics[width=\linewidth]{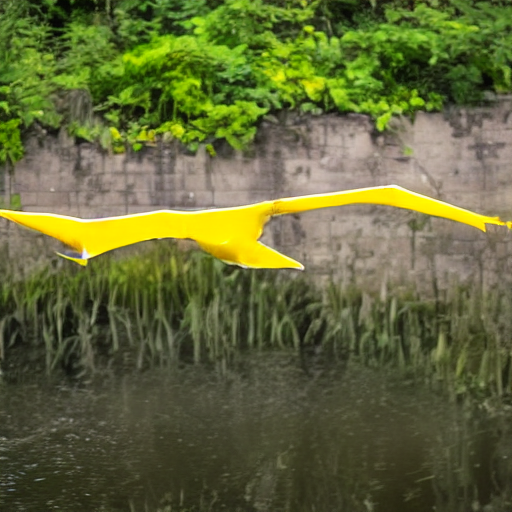}
   \caption{Ours}
\end{subfigure}%
\begin{subfigure}[b]{0.25\linewidth}
        \includegraphics[width=\linewidth]{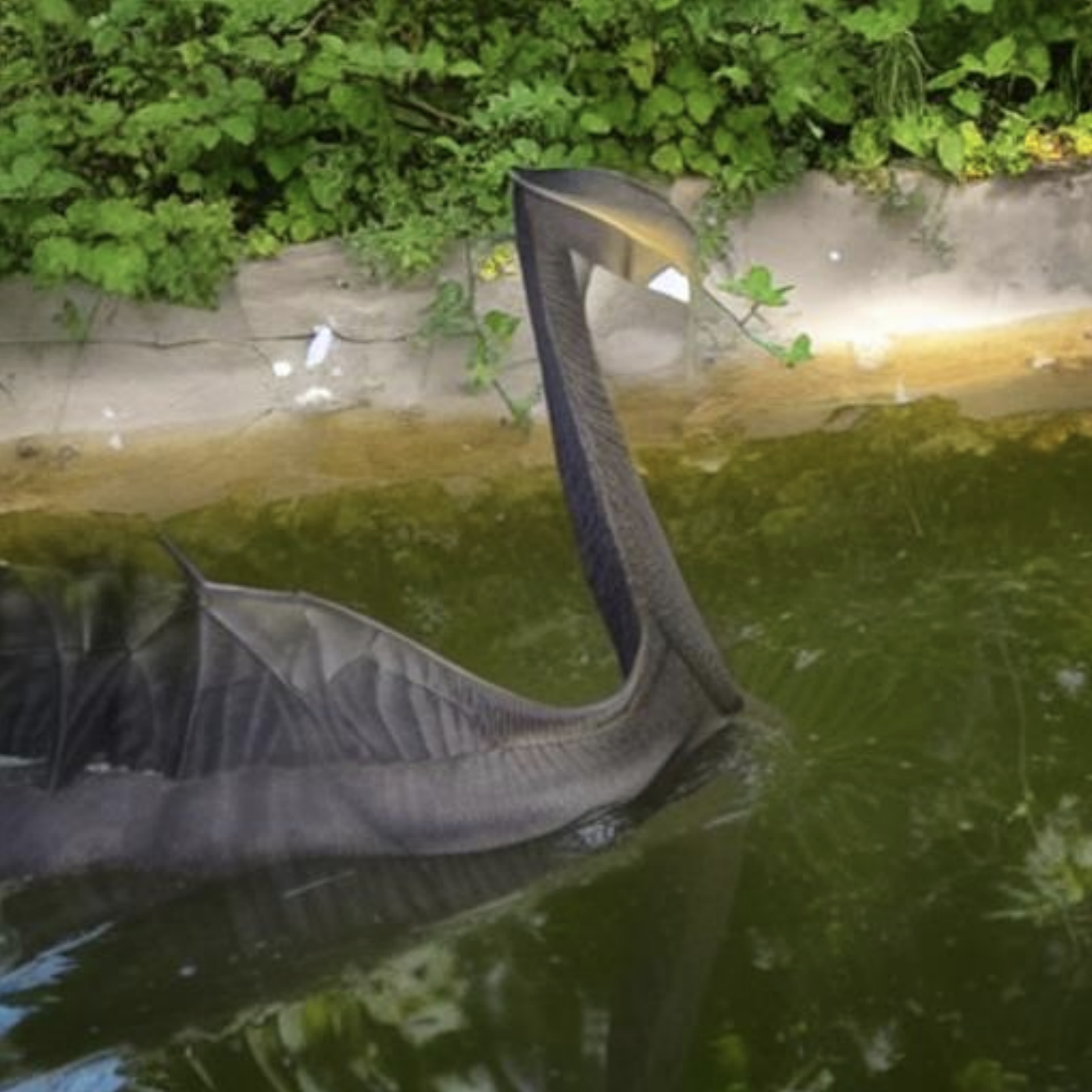}\\
        \includegraphics[width=\linewidth]{fateterosaur.png}
   \caption{FateZero}
\end{subfigure}%
\begin{subfigure}[b]{0.25\linewidth}
        \includegraphics[width=\linewidth]{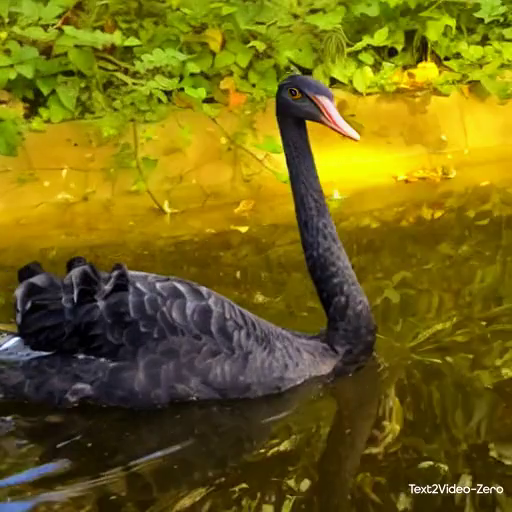}\\
        \includegraphics[width=\linewidth]{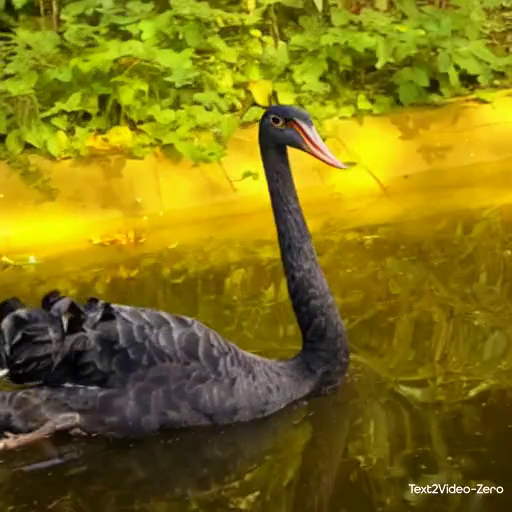}
   \caption{T2V-Zero}
\end{subfigure}
    \vspace{-6mm}
\caption{\textbf{Qualitative comparison} of our method with FateZero and Text2Video-Zero (T2V-Zero). Best viewed with zoom-in}
\label{fig:qualityfateT2V}
\end{figure}
\begin{figure}
\centering
\begin{subfigure}[b]{0.33\linewidth}
   \includegraphics[width=\linewidth]{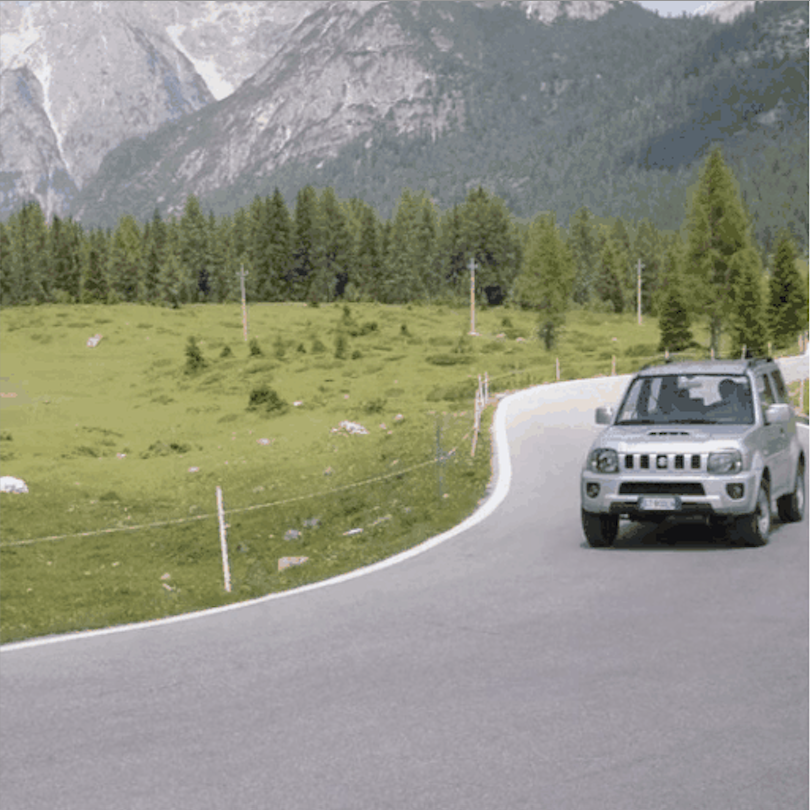}\\
    \includegraphics[width=\linewidth]{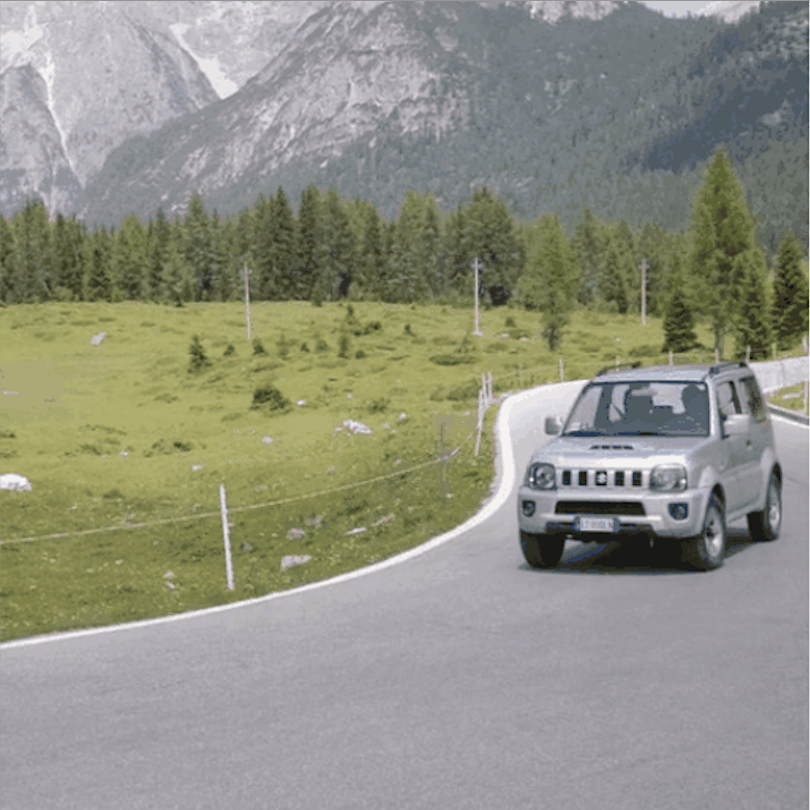}
   \caption{Input}
\end{subfigure}%
\begin{subfigure}[b]{0.33\linewidth}
        \includegraphics[width=\linewidth]{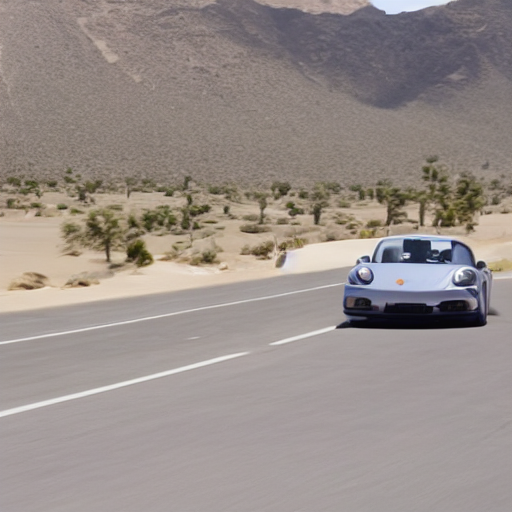}\\
        \includegraphics[width=\linewidth]{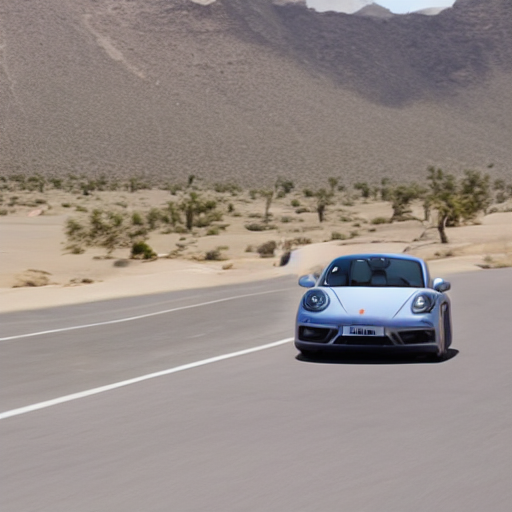}
   \caption{Ours}
\end{subfigure}%
\begin{subfigure}[b]{0.33\linewidth}
        \includegraphics[width=\linewidth]{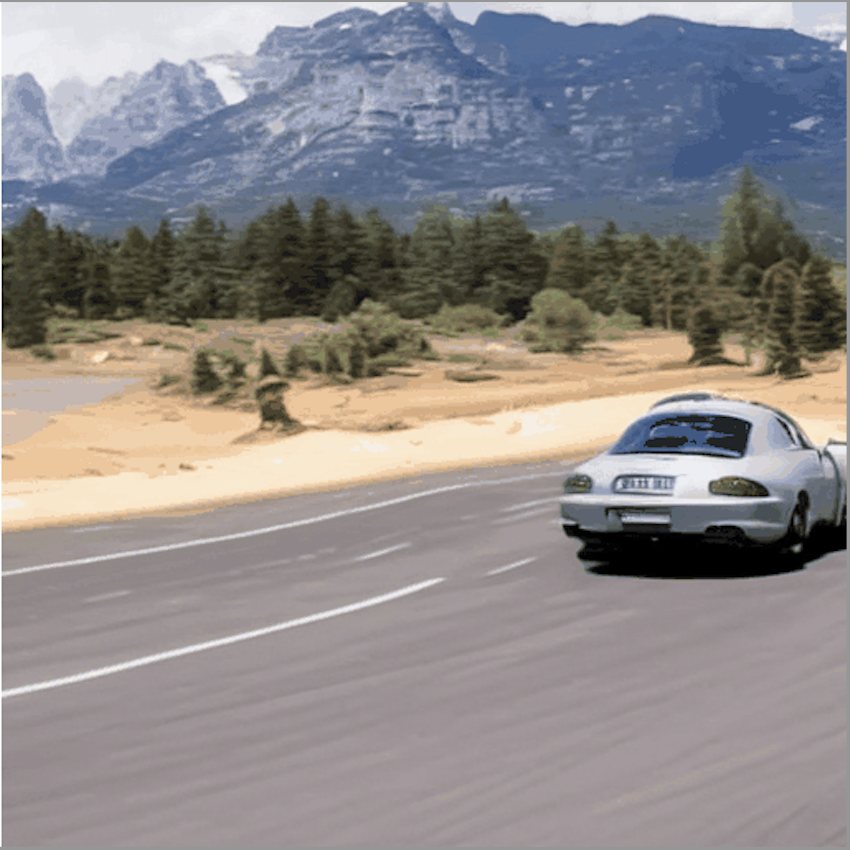}\\
        \includegraphics[width=\linewidth]{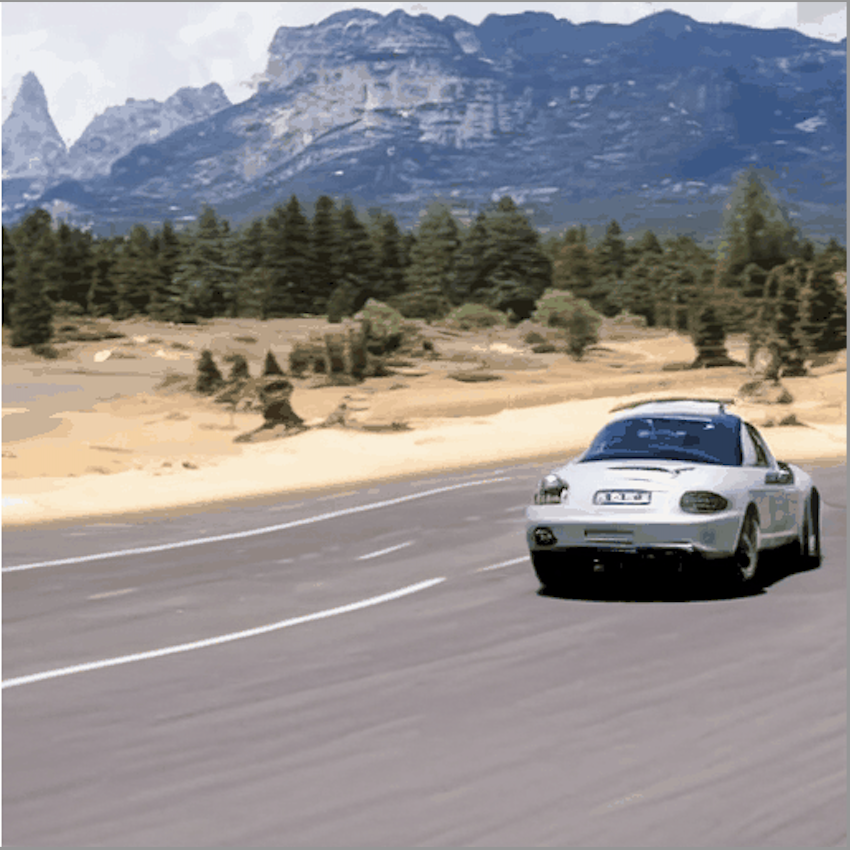}
   \caption{vid2vid-zero}
\end{subfigure}
   \vspace{-3mm}
\caption{\textbf{Qualitative comparison} of our method with vid2vid-zero. Best viewed with zoom-in}
\label{fig:qualityvid}
\end{figure}
\begin{figure}
\centering
\begin{subfigure}[b]{0.25\linewidth}
   \includegraphics[width=\linewidth]{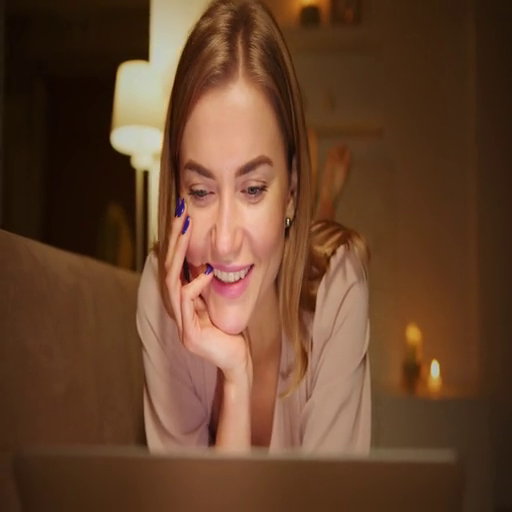}\\
    \includegraphics[width=\linewidth]{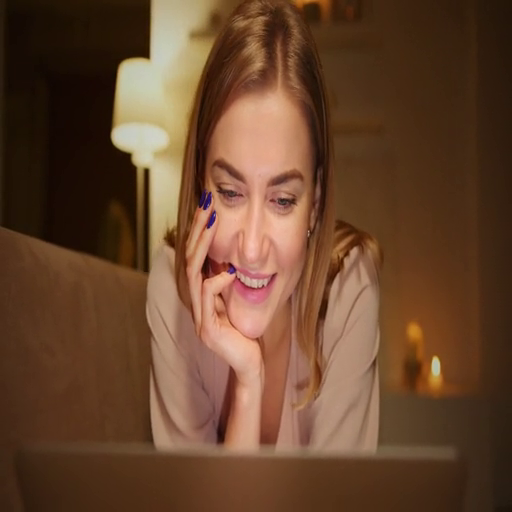}
   \caption{Input}
\end{subfigure}%
\begin{subfigure}[b]{0.25\linewidth}
        \includegraphics[width=\linewidth]{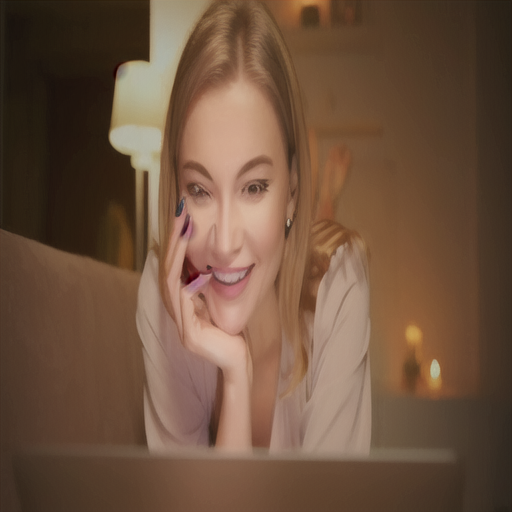}\\
        \includegraphics[width=\linewidth]{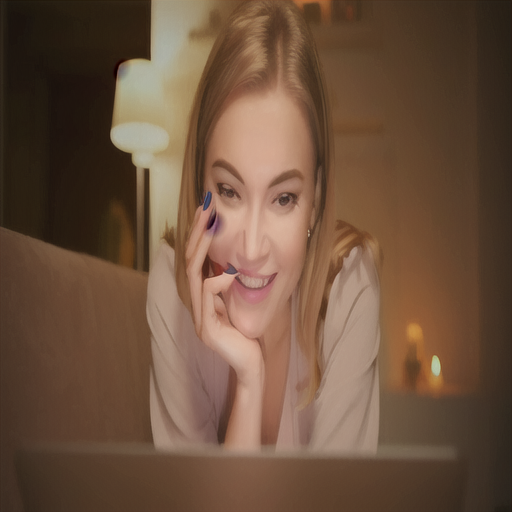}
   \caption{CG Style}
\end{subfigure}%
\begin{subfigure}[b]{0.25\linewidth}
        \includegraphics[width=\linewidth]{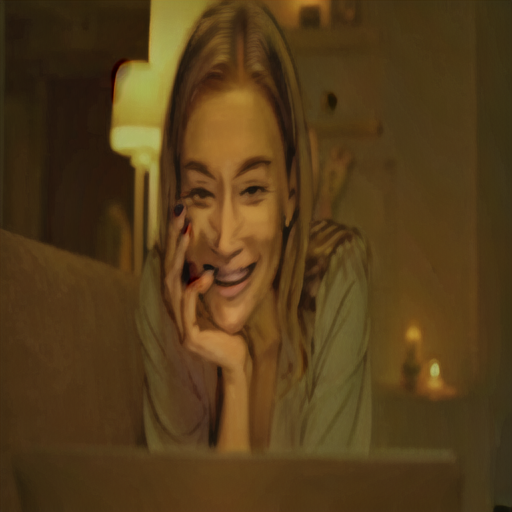}\\
        \includegraphics[width=\linewidth]{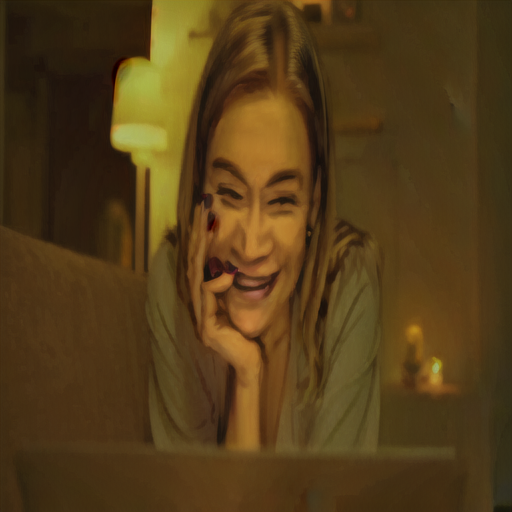}
   \caption{Ghibli Cartoon}
\end{subfigure}%
\begin{subfigure}[b]{0.25\linewidth}
        \includegraphics[width=\linewidth]{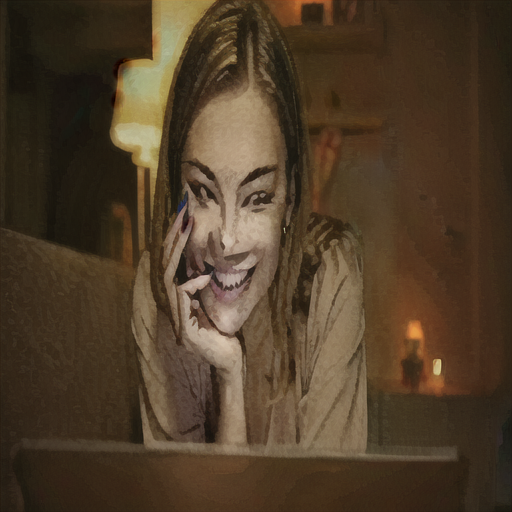}\\
        \includegraphics[width=\linewidth]{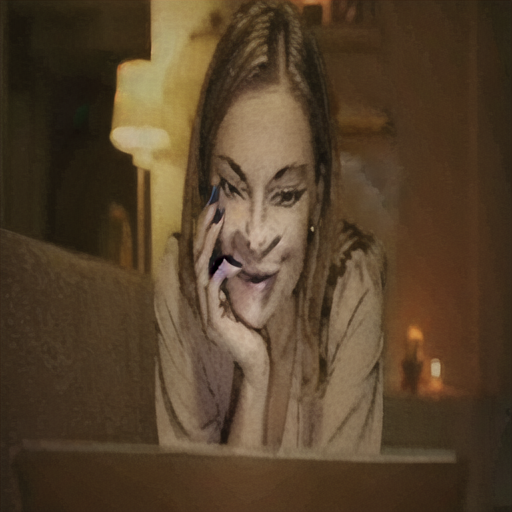}
   \caption{Ink Painting}
\end{subfigure}
\caption{\textbf{More Results}: Demonstration of editing for cartoon and painting styles}
\vspace{-6mm}
\label{fig:more_res1}
\end{figure}

\begin{figure}
\centering
\begin{subfigure}[b]{\linewidth}
   \includegraphics[width=0.25\linewidth]{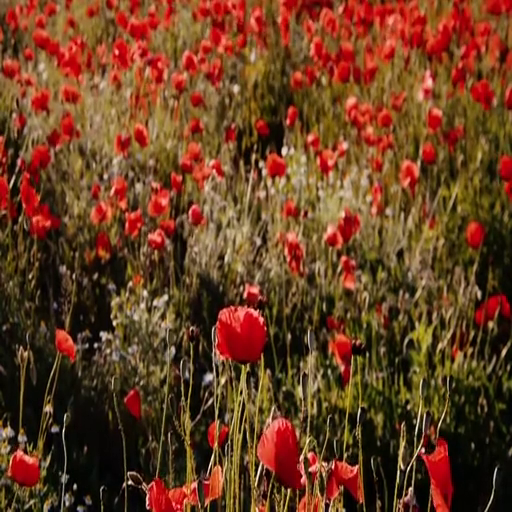}%
    \includegraphics[width=0.25\linewidth]{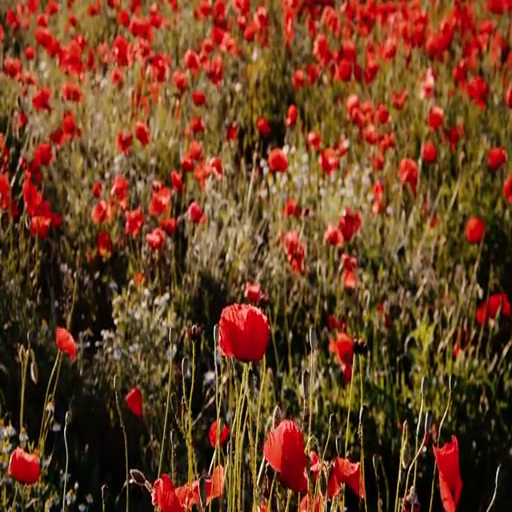}%
    \includegraphics[width=0.25\linewidth]{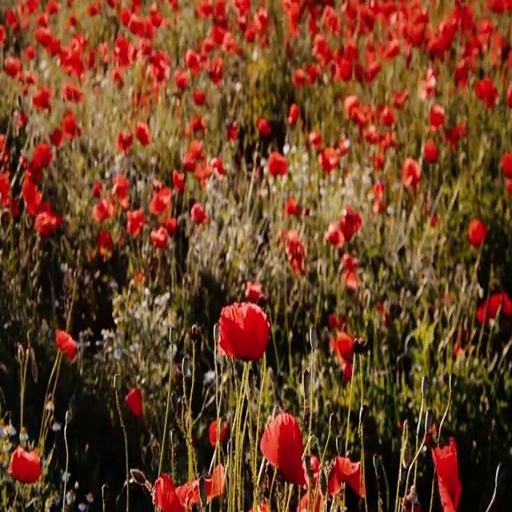}%
    \includegraphics[width=0.25\linewidth]{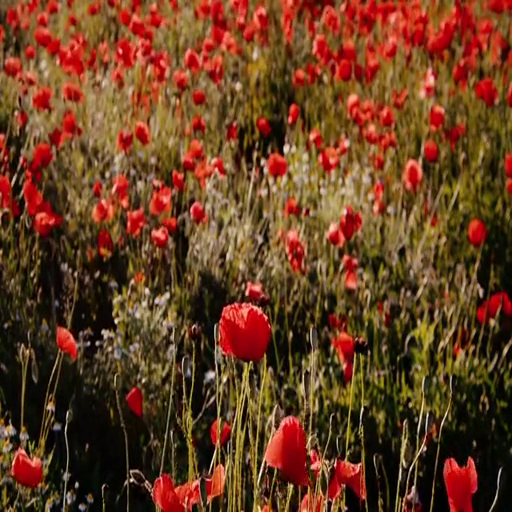}\\
    \includegraphics[width=0.25\linewidth]{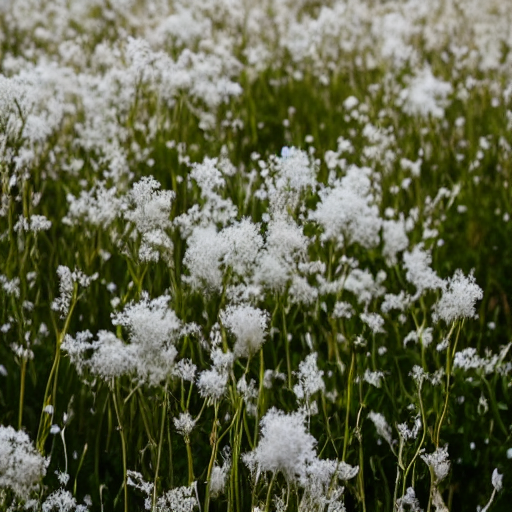}%
    \includegraphics[width=0.25\linewidth]{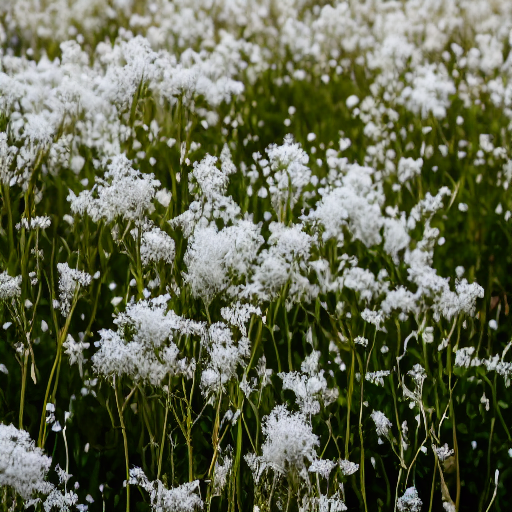}%
    \includegraphics[width=0.25\linewidth]{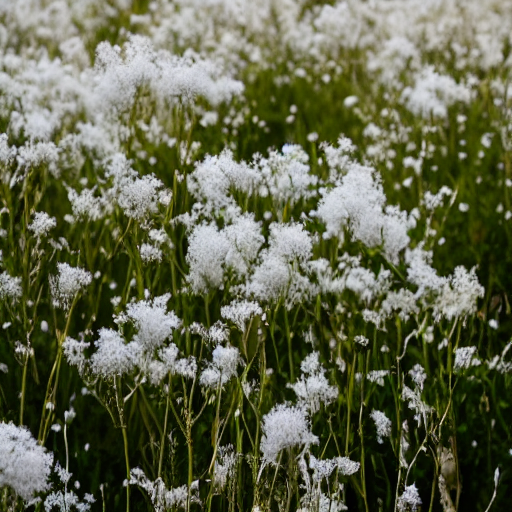}%
    \includegraphics[width=0.25\linewidth]{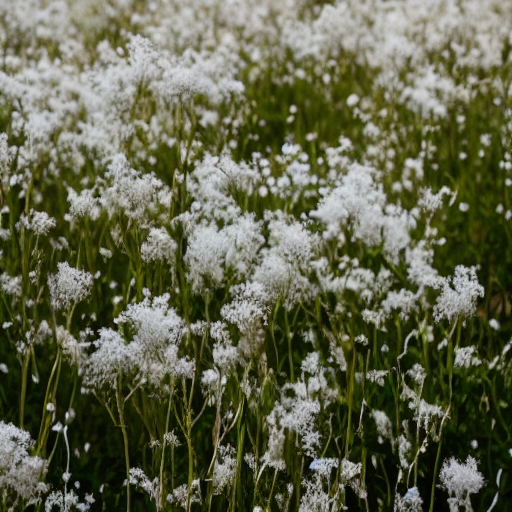}
   \caption{\small\textit{{\textbf{Source Prompt}: Blooming field of \textcolor{red}{red poppy flowers}. \textbf{Target Prompt}: Blooming field of \textcolor{cyan}{white poppy flowers}}}}
   \label{fig:poppy} 
\end{subfigure}%
\vspace{1mm}
\begin{subfigure}[b]{\linewidth}
        \includegraphics[width=0.25\linewidth]{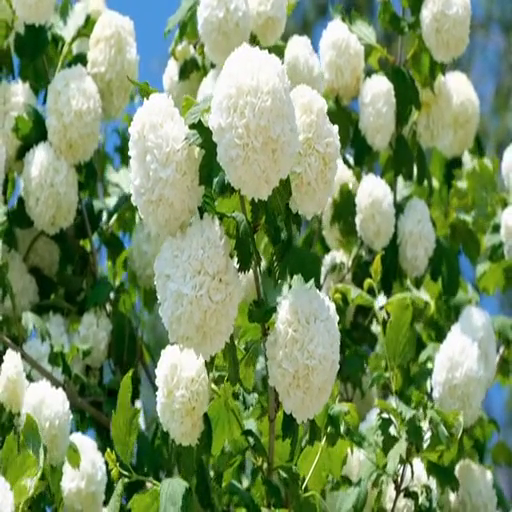}%
        \includegraphics[width=0.25\linewidth]{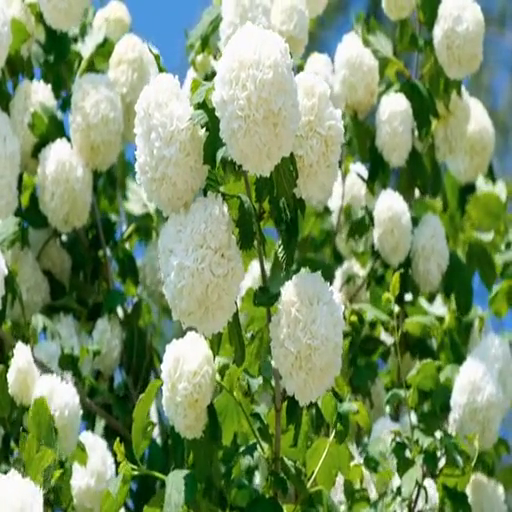}%
        \includegraphics[width=0.25\linewidth]{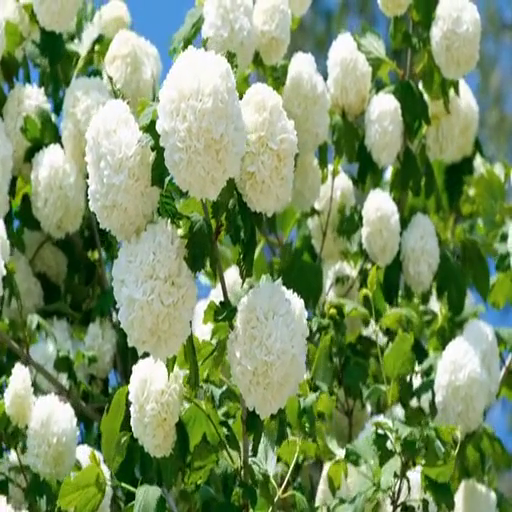}%
        \includegraphics[width=0.25\linewidth]{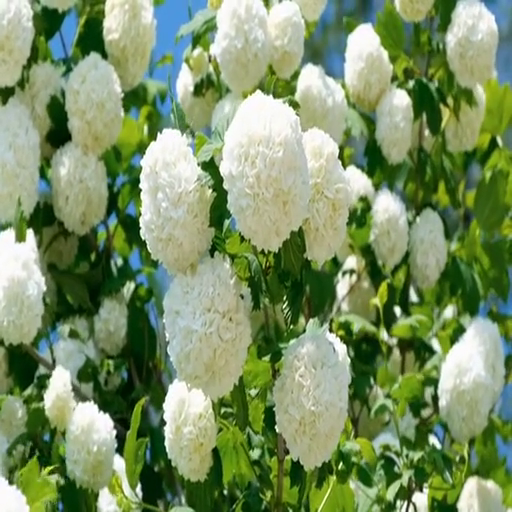}\\
        \includegraphics[width=0.25\linewidth]{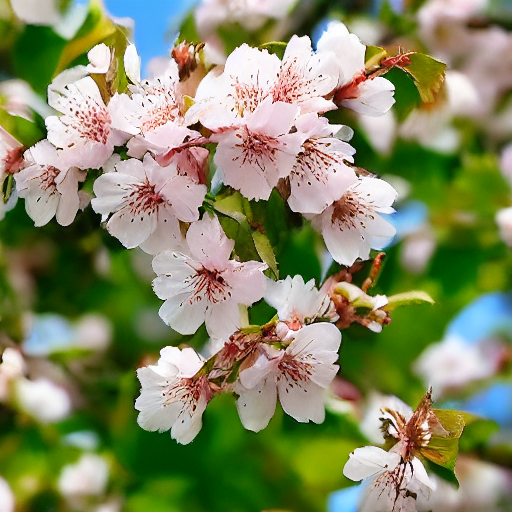}%
        \includegraphics[width=0.25\linewidth]{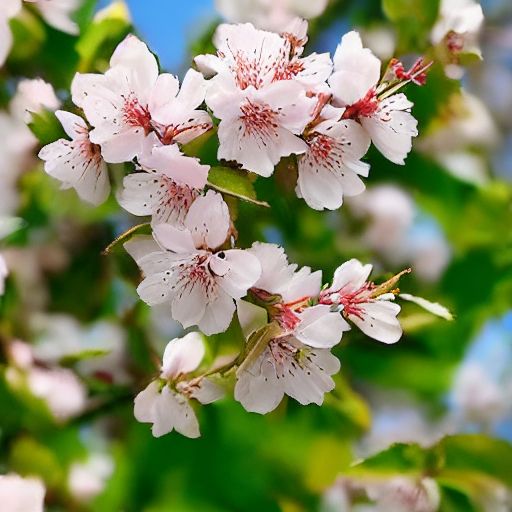}%
        \includegraphics[width=0.25\linewidth]{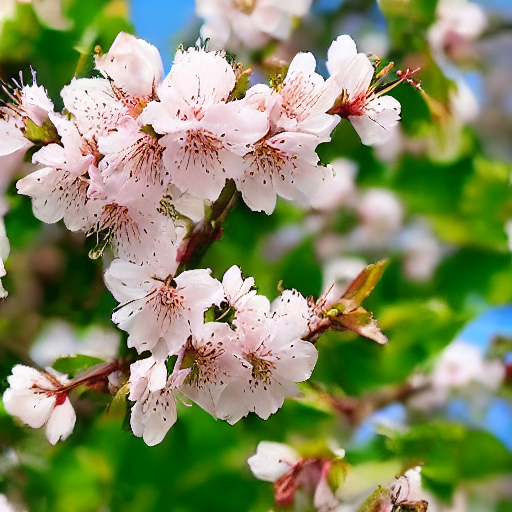}%
        \includegraphics[width=0.25\linewidth]{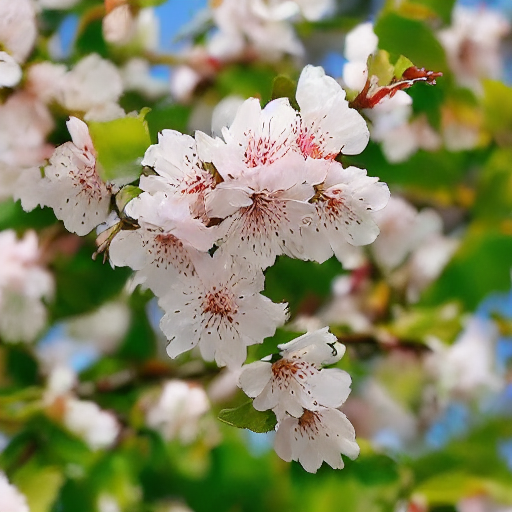}
   \caption{\small\textit{{\textbf{Source Prompt}: {White Snowball} Flowers. \textbf{Target Prompt}: \textcolor{Rhodamine}{Cherry Blossom} Flowers}}}
   \label{fig:snowball}
\end{subfigure}%
\vspace{1mm}
\begin{subfigure}[b]{\linewidth}
        \includegraphics[width=0.25\linewidth]{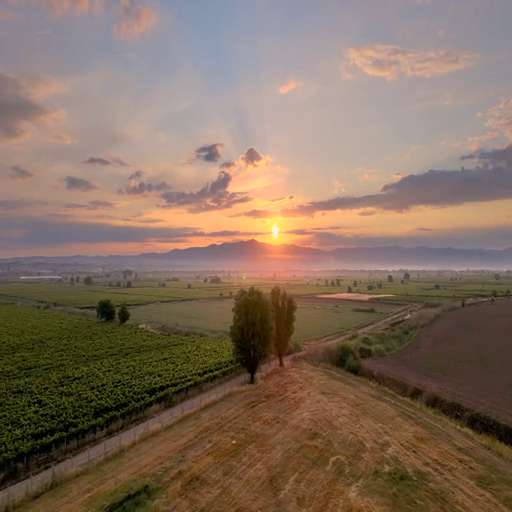}%
        \includegraphics[width=0.25\linewidth]{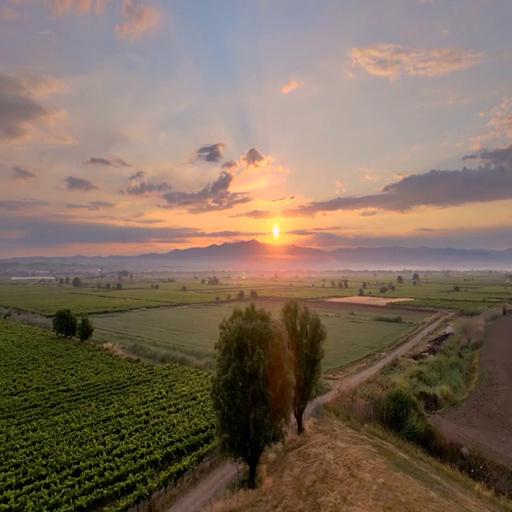}%
        \includegraphics[width=0.25\linewidth]{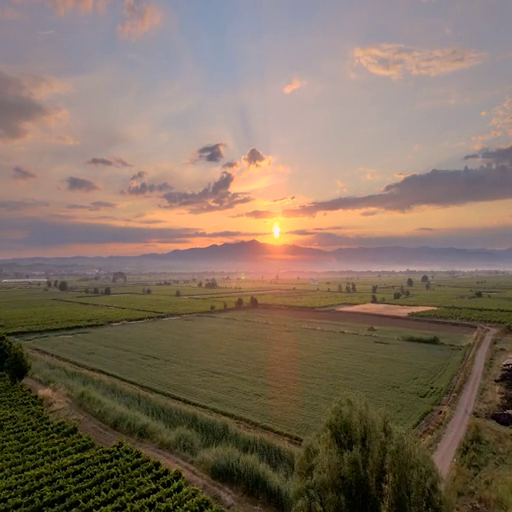}%
        \includegraphics[width=0.25\linewidth]{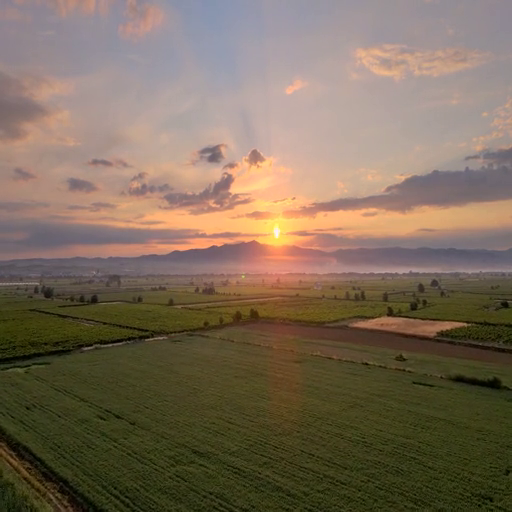}\\
        \includegraphics[width=0.25\linewidth]{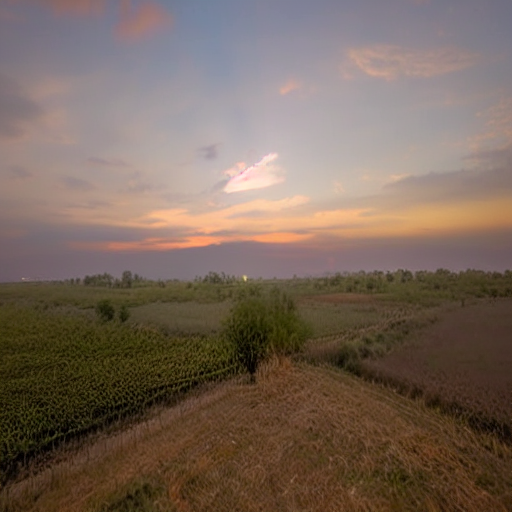}%
        \includegraphics[width=0.25\linewidth]{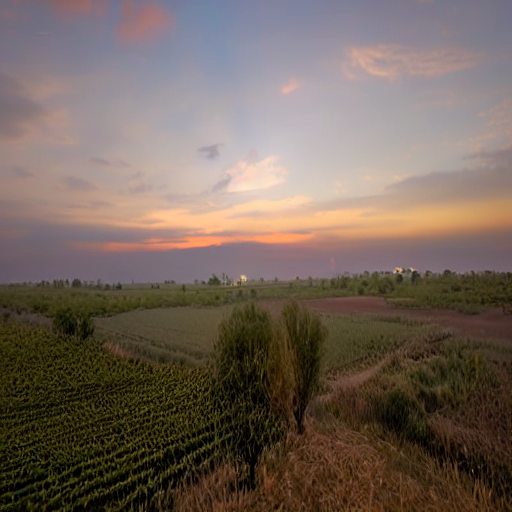}%
        \includegraphics[width=0.25\linewidth]{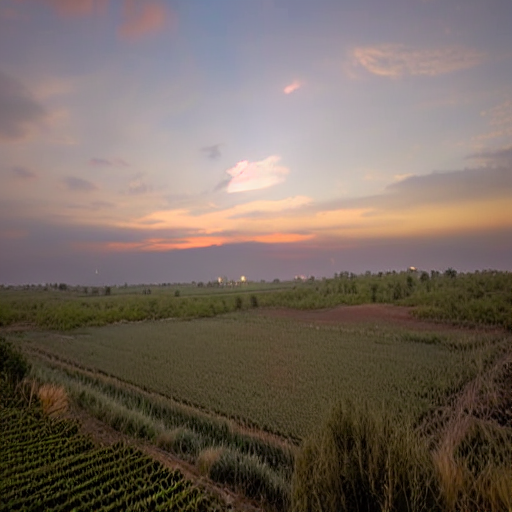}%
        \includegraphics[width=0.25\linewidth]{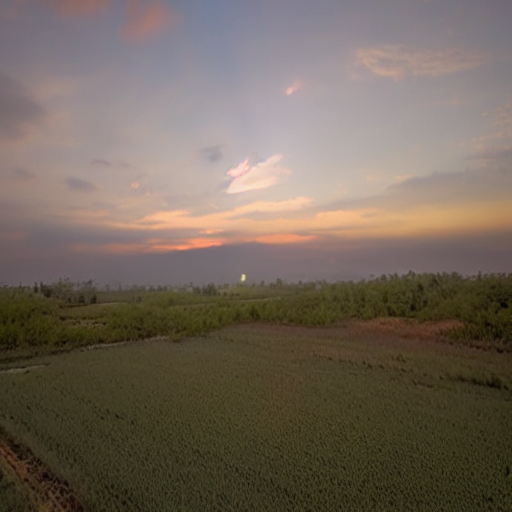}
   \caption{\small\textit{{\textbf{Source Prompt}: \textbf{Morning view} over a farm. \textbf{Target Prompt}: \textcolor{orange}{Sunset view} over a farm}}}
   \label{fig:sunset}
\vspace{-3mm}
\end{subfigure}
\caption{\textbf{More Results}: Demonstration of editing over fine-grained structure, shape and color}
\label{fig:more_res2}
\vspace{-5mm}
\end{figure}
\subsection{Comparison with State-of-the-Art Methods}
We compare our approach with four recent zero-shot methods, namely: FateZero \cite{qi2023fatezero}, vid2vid-zero \cite{wang2023zero}, Pix2Video \cite{ceylan2023pix2video}, and Text2Video-Zero \cite{khachatryan2023text2video}, where the code for all the baselines is publicly available, but they generate plausible results only for editing templates mentioned in their code base. In particular, vid2vid-zero generated frames totally different from the input. For the sake of comparison, we have taken most of the videos that are common among these baselines. As shown in Figure \ref{fig:qualityfateT2V}, the case of editing "\textit{swan}" to "\textit{\textcolor{Goldenrod}{yellow terosaur}}", this is mentioned by Fatezero as their limitation in their paper. As shown, we are able to successfully replace the "\textit{swan}" to "\textit{\textcolor{Goldenrod}{yellow terosaur}}". Additionally, we compared this with Text2Video-Zero as well since this uses the ControlNet, hence, it has added information to edit the videos, but it generates frames that are not even close to the target prompt, instead, it changes the colour of the wall to yellow. Similarly, in Figure \ref{fig:qualityvid}, we have compared our method against vid2vid-zero for the input video and target prompt mentioned in their zero shot results. The generated frames, as per their code base, show the backside of the "\textit{Porsche car}" while moving forward, whereas in the input video, the Jeep is moving forward and is front-facing. We showed that our results are much better in terms of resemblance to input and target prompts. Following \cite{qi2023fatezero, zhao2023controlvideo, esser2023structure, wu2022tune} we also conducted the quantitative evaluation using the trained CLIP \cite{radford2021learning} model. Specifically, we show the "\textbf{Temporal}" \cite{esser2023structure} to measure the temporal consistency between consecutive frames by measuring cosine similarity between all pairs of consecutive frames. Another measure, "\textbf{Edit Acc}" \cite{radford2021learning, wu2022tune} to measure the editing accuracy in the generated frames, is calculated as the percentage of generated frames having a higher similarity for the target prompt than the source prompt. Additionally, we evaluated our method on two user studies metrics ('\textbf{Edit}' and '\textbf{Temporal}') are measured to measure the editing quality of our system from the perspective of the applicability of our method in terms of usage in a real environment. Specifically, we measured the rank of our proposed method for temporal consistency ('\textbf{Temporal}') across the frames in edited video and overall frame-wise editing ('\textbf{Edit}') for a given target prompt. We asked 20 subjects to rank the editing method, with nine sets of comparisons in each study. As shown in Table \ref{tab:quant}, our proposed method \textsc{Infusion} outperforms for all the CLIP metrics, hence achieving the best temporal consistence and better per-frame editing accuracy. Moreover, our method is truly zero shot since we have not used any other pre-trained diffusion other than Stable Diffusion v1.5\cite{rombach2022high} as opposed to \textbf{Fatezero}, which uses the one-shot trained model for the target prompt "\textit{A Porsche car driving down a curvy road in the countryside}". Apart from CLIP metrics, our method is more reliable to put into real world editing since it earns user preferences the best among all methods in both (\textbf{Edit} and \textbf{Temporal}) aspects.
\begin{table}[]
\centering
\small
\resizebox{\columnwidth}{!}{%
\begin{tabular}{lcccc}
\toprule
\multicolumn{1}{c}{\multirow{2}{*}{Method}} & \multicolumn{2}{c}{CLIP Metrics $\uparrow$} & \multicolumn{2}{c}{User Study $\downarrow$} \\ \cmidrule{2-5} 
\multicolumn{1}{c}{}                        & Temporal             & Edit Acc             & Temporal             & Edit                 \\ \midrule
Tune-A-Video                                & 0.934                & 0.738                & 2.79                 & 2.73                 \\
vid2vid-zero                                & 0.951                & 0.696                & 2.71                     &  2.69                 \\
FateZero                                    & 0.954                & 0.894                &  1.89                   &   2.52                  \\
Pix2Video                                   & 0.912                & 0.701                &  2.60                    &  1.98                    \\
Text2Video-Zero                             & 0.959                & 0.902                &   1.81                   &  1.77                    \\ \midrule
\textbf{Ours}                               & \textbf{0.971}       & \textbf{0.915}       & \textbf{1.78}            & \textbf{1.31}            \\ \bottomrule
\end{tabular}%
}
\caption{\textbf{Quantitative evaluation}: For both user study and CLIP metrics \textsc{Infusion} outperforms all the baselines in terms of temporal consistency and per frame editing accuracy}
\label{tab:quant}
\vspace{-8mm}
\end{table}

\begin{figure}
\centering
\begin{subfigure}[b]{0.2\linewidth}
    \includegraphics[width=\linewidth]{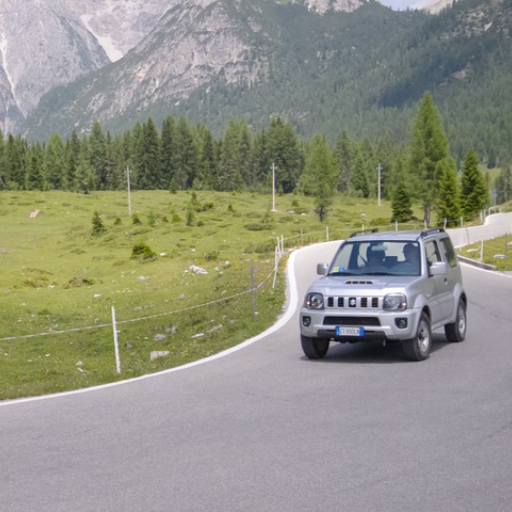}\\
   \includegraphics[width=\linewidth]{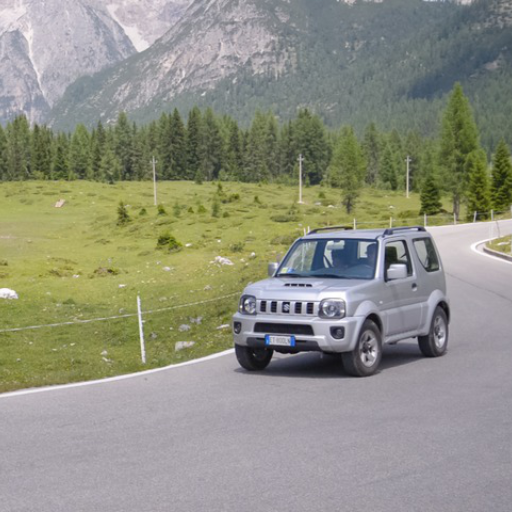}\\
    \includegraphics[width=\linewidth]{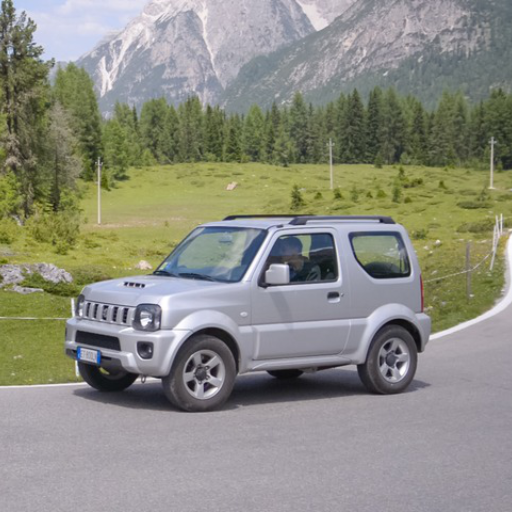}
   \caption*{Input}
\end{subfigure}%
\begin{subfigure}[b]{0.2\linewidth}
        \includegraphics[width=\linewidth]{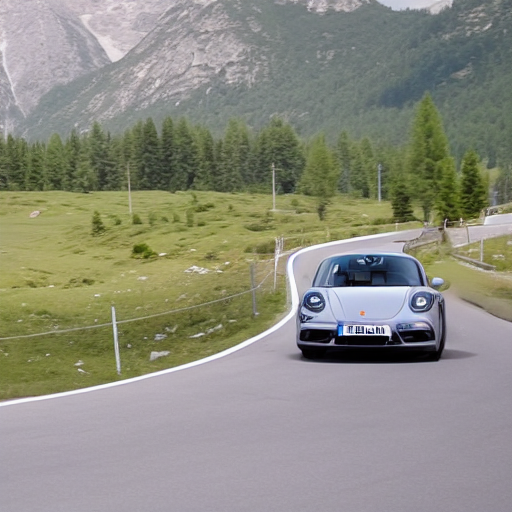}\\
        \includegraphics[width=\linewidth]{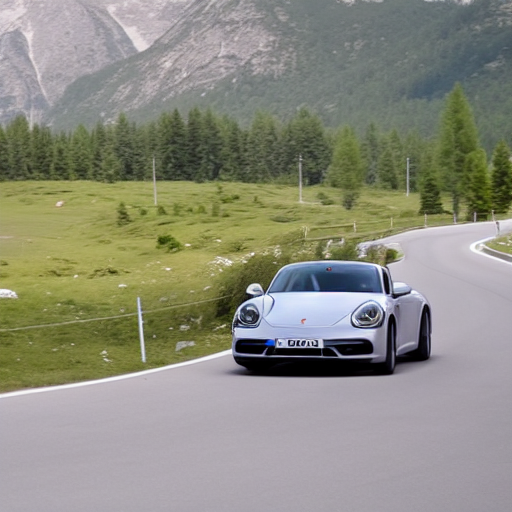}\\
        \includegraphics[width=\linewidth]{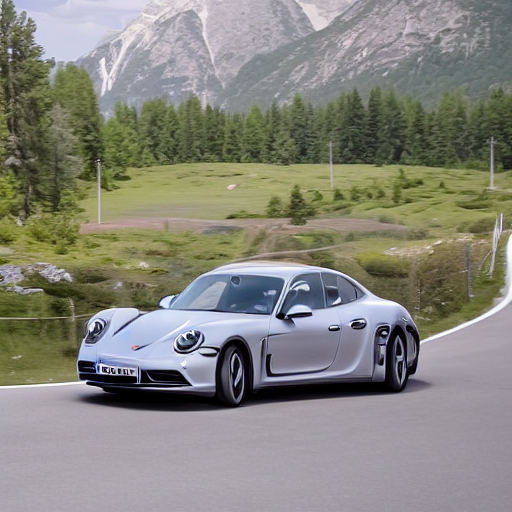}
   \caption*{Ours}
\end{subfigure}%
\begin{subfigure}[b]{0.2\linewidth}
        \includegraphics[width=\linewidth]{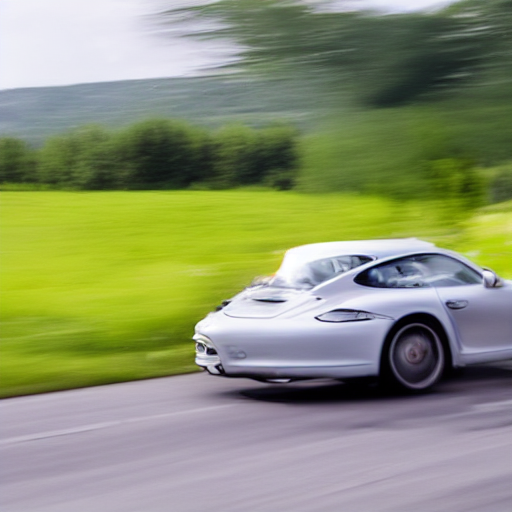}\\
        \includegraphics[width=\linewidth]{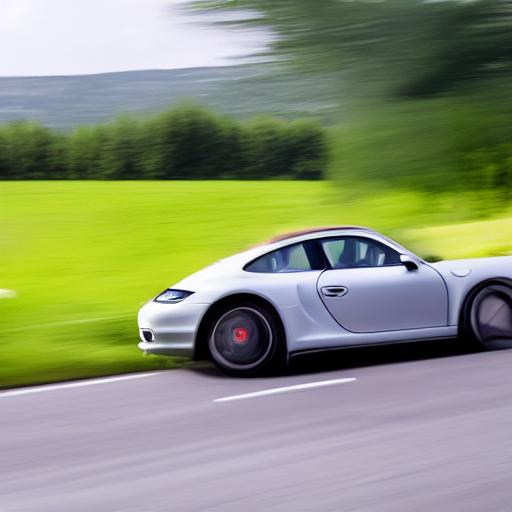}\\
        \includegraphics[width=\linewidth]{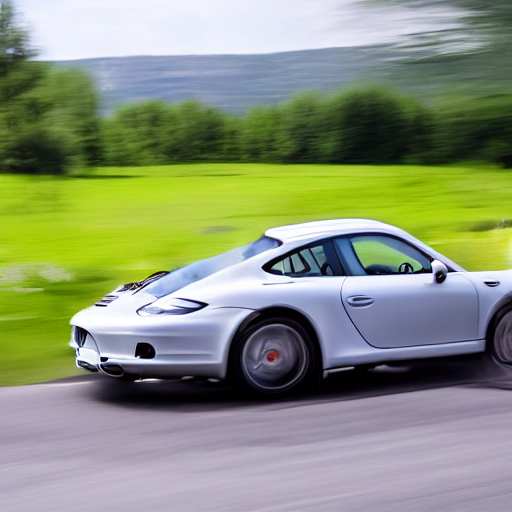}
   \caption*{w/o $f_t^l$-inject}
\end{subfigure}%
\begin{subfigure}[b]{0.2\linewidth}
        \includegraphics[width=\linewidth]{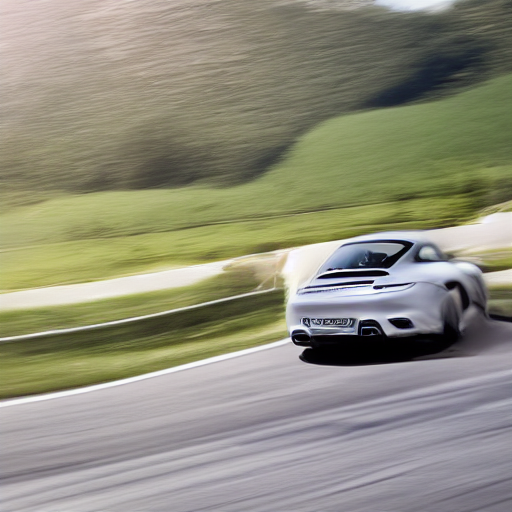}\\
        \includegraphics[width=\linewidth]{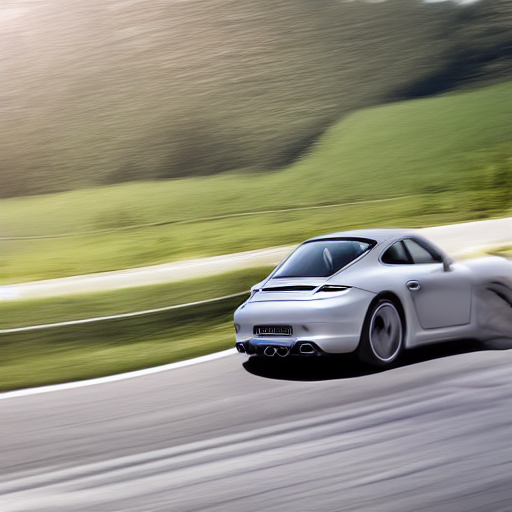}\\
        \includegraphics[width=\linewidth]{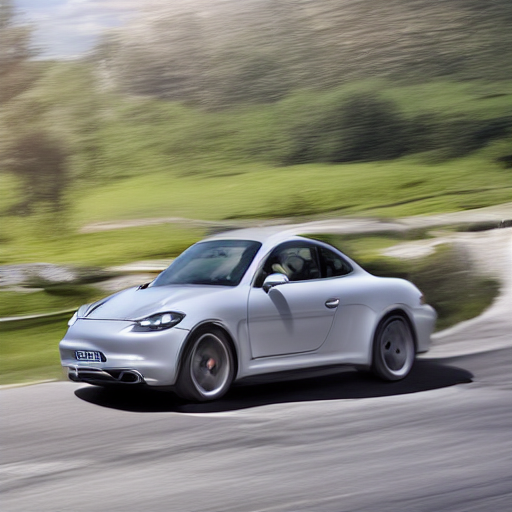}
   \caption*{w/o att-inject}
\end{subfigure}%
\begin{subfigure}[b]{0.2\linewidth}
        \includegraphics[width=\linewidth]{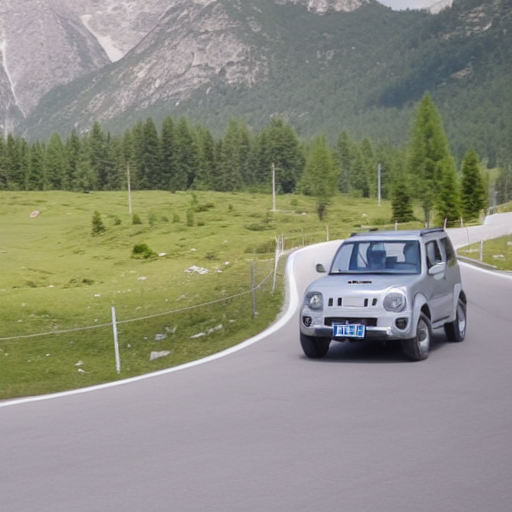}\\
        \includegraphics[width=\linewidth]{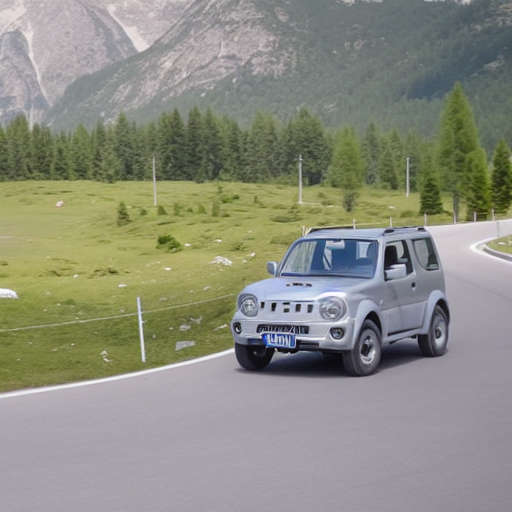}\\
        \includegraphics[width=\linewidth]{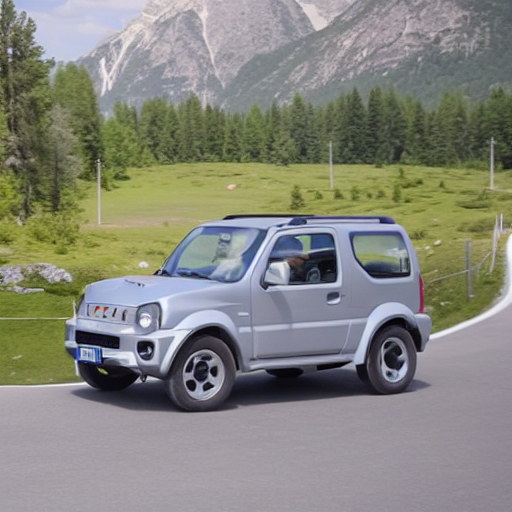}
   \caption*{w/o \textsc{Fusion}}
\end{subfigure}
\caption{\textbf{Ablation Study} of Feature Injection, Attention Injection and Attention Fusion. Prompts used are $P_s$: "\textit{A \textbf{silver jeep} driving down a curvy road in the countryside}", $P_e$: "\textit{A \textcolor{red}{Porsche car} driving down a curvy road in the countryside}". Without any of these components the foreground or background or both details are missing from the edited video.}
\label{fig:ablat}
\vspace{-6mm}
\end{figure}
\subsection{Ablation Study}
Despite proving the effectiveness of \textsc{Infusion}, in this section we will present the ablation study (shown in Figure \ref{fig:ablat}) of various components in our editing method to discuss the importance of each and their contribution towards the edited video.\\
\textbf{\textsc{Injection}} is studied in Figure \ref{fig:ablat} as shown in the third and fourth columns. The column named "\textbf{\textit{w/o $f_t^{l}$-inject}}" is the ablation study of feature injection, it depicts the complete change in frames including both foreground (\textit{Porsche car}) and background (\textit{curvy road and countryside as in source frames}). Though, we wanted to change the structure of "\textit{jeep} $\rightarrow$ \textit{Porsche car}", but without feature injection it changes the complete source layout/background as well, which is not a desirable change. The column named "\textbf{\textit{w/o att-inject}}" is the ablation for attention injection. As shown, without attention injection, the rest of the background is faded, however, due to feature injection, the source layout/background are retained ("\textit{curvy road and countryside}"). Additionally, due to feature injection, the target concept is highlighted ("\textit{Porsche car}") but the remaining source layout is faded, and hence it is very much required to fill the source concepts on the highlighted target concepts, which is proposed to be done with self-attention injection since it can inject the source concepts while keeping the importance of highlighted target concepts injected using differential features.\\
\textbf{\textsc{Attention Fusion}} is also studied in Figure \ref{fig:ablat}, where we have removed self attention fusion (shown in the fifth column) as discussed in equation \ref{selfusion} but cross attention mixing is present. The resulting frames show little or no change, with some shape distortion in the front part of the Jeep, as if the diffusion is trying to align the Jeep structure with that of the car. As expected, the change injected using differential feature and attention until $S_2$ steps is wiped, and the concepts that are highlighted also get wiped due to continued diffusion steps without masking, and hence it will generate a structure similar to the source.
\subsection{\textsc{\textbf{More Results}: Flexible Structure, Color and Style}}
Our proposed method has shown decent results in editing the structure, colour, and shape at finer details (Figures \ref{fig:more_res2} and \ref{fig:more_res1}) with source content preserved. Specifically, as shown in Figure \ref{fig:poppy} the source content contains lot of cluttered red flowers, which are edited to white flowers (fine-grained colour editing) with decent temporal consistency and accuracy over the frames. Similarly, Figure \ref{fig:snowball} contains a lot of snowball flowers, which are edited to "\textit{cherry blossom}" flowers (fine-grained structure and shape editing) at all viewing angles demonstrated over frames. Figure \ref{fig:sunset} demonstrated fine-grained structure editing from "\textit{morning}" view to "\textit{sunset}" view with just deleting the sun and associated rays over the farm and remaining everything intact. Other results include style editing ranging from cartoon to painting styles, as shown in Figure \ref{fig:more_res1}. Here, as we see, the details like nails, pose, etc. of the woman are preserved. 
More results will be added to the github page: \href{https://infusion-zero-edit.github.io/}{\textcolor{RubineRed}{\textit{https://infusion-zero-edit.github.io/}}}
\section*{Conclusion} 
In this work, we have presented the generalised zero-shot text-based video editing framework with no additional models like ControlNet\cite{zhang2023adding} and with no training or fine-tuning of the pre-trained image diffusion model. Only pre-trained Stable Diffusion v1.5\cite{rombach2022high} is used for all the edited videos without the need for any customised image diffusion.

{\small
\bibliographystyle{ieee_fullname}
\bibliography{egpaper_final}
}
\end{document}